%% file: main.tex
\documentclass{article} 
\usepackage{iclr2027_conference,times}

\input{math_commands.tex}

\usepackage{hyperref}
\hypersetup{hidelinks} 
\usepackage{url}
\usepackage{graphicx}
\usepackage{booktabs}
\usepackage{multirow}
\usepackage{placeins}
\usepackage{longtable}[=2020-01-07]
\usepackage{array}[=2024-06-01]
\usepackage[table]{xcolor}
\definecolor{oursrow}{HTML}{EAF4F2}
\usepackage{amsmath}
\usepackage{amssymb}
\usepackage{etoolbox}
\renewcommand{\eqref}[1]{(\ref{#1})}
\newcommand{\papertablesetup}{%
  \small\rmfamily
  \setlength{\tabcolsep}{3pt}%
  \renewcommand{\arraystretch}{1.12}}


\newcommand{\compactfigure}{%
  \setlength{\parskip}{0pt}%
  \setlength{\abovecaptionskip}{2pt}%
  \setlength{\belowcaptionskip}{0pt}}
\newcommand{\compacttable}{%
  \setlength{\parskip}{0pt}%
  \setlength{\abovecaptionskip}{0pt}%
  \setlength{\belowcaptionskip}{2pt}}
\AddToHook{env/figure/begin}{\compactfigure}
\AddToHook{env/figure*/begin}{\compactfigure}
\AddToHook{env/table/begin}{\compacttable}
\AddToHook{env/table*/begin}{\compacttable}

\makeatletter
\newcommand{\maintextspacing}{%
  \setlength{\parskip}{2pt plus 0.5pt minus 0.5pt}%
  \setlength{\abovedisplayskip}{4pt plus 1pt minus 1pt}%
  \setlength{\belowdisplayskip}{4pt plus 1pt minus 1pt}%
  \setlength{\abovedisplayshortskip}{0pt plus 1pt}%
  \setlength{\belowdisplayshortskip}{3pt plus 1pt minus 1pt}%
  \renewcommand{\section}{\@startsection{section}{1}{\z@}%
    {-6pt plus -1pt minus -1pt}{4pt plus 1pt}{\large\sc\raggedright}}%
  \renewcommand{\subsection}{\@startsection{subsection}{2}{\z@}%
    {-5pt plus -1pt minus -1pt}{3pt plus 1pt}{\normalsize\sc\raggedright}}%
  \renewcommand{\paragraph}{\@startsection{paragraph}{4}{\z@}%
    {3pt plus 1pt minus 1pt}{-1em}{\normalsize\bf}}}
\makeatother

\title{\normalfont Spatial Latent Reasoning for\\Embodied Reference Understanding}

\author{%
\normalfont Ling Li\textsuperscript{1}, Jianhui Zhong\textsuperscript{2}, Wei Liu\textsuperscript{2}, Zheng Jiang\textsuperscript{1},\\
\normalfont Yuxuan Liu\textsuperscript{1}, Jingyu Li\textsuperscript{3}, Zhidong Deng\textsuperscript{1}\\[2pt]
\normalfont\small \textsuperscript{1}Tsinghua University\quad \textsuperscript{2}Dalian University of Technology\\
\normalfont\small \textsuperscript{3}University of Science and Technology of China}
\hypersetup{%
  pdftitle={Spatial Latent Reasoning for Embodied Reference Understanding},
  pdfauthor={Ling Li, Jianhui Zhong, Wei Liu, Zheng Jiang, Yuxuan Liu, Jingyu Li, Zhidong Deng}}

\makeatletter
\patchcmd{\@maketitle}{\rule{\z@}{24pt}\@author}{\rule{\z@}{14pt}\@author}{}{\errmessage{Author spacing patch failed}}
\patchcmd{\@maketitle}{\vskip 0.3in minus 0.1in}{\vskip 6pt}{}{\errmessage{Title spacing patch failed}}
\makeatother
\iclrfinalcopy 
\begin{document}

\maketitle
\lhead{} 
\renewcommand{\headrulewidth}{0pt}
\raggedbottom 
\suppressfloats[t] 

\begin{abstract}
Pointing-gesture visual grounding requires connecting hand geometry with the visual identity and extent of a referred object. A central challenge for continuous latent reasoning is how to organize these complementary cues into useful intermediate supervision. We propose \emph{Spatial Latent Reasoning (SLR)}, a framework that structures this supervision around an ordered sequence of geometric and visual states. A spatial ray state is supervised by fingertip position and pointing direction, followed by four states aligned with target-region features. To construct the visual targets, we introduce parity pooling, which applies polyphase grouping to average region tokens on four interleaved spatial supports. All states are generated recurrently during training and inference; auxiliary annotations are required only during training. On EgoPoint-Ground, the framework improves mIoU over same-backbone supervised fine-tuning by \textbf{2.8, 17.5, and 21.1 percentage points} on Qwen3.5-4B, Qwen2.5-VL-7B, and Qwen3-VL-8B, respectively, with improvements on both hard subsets. On YouRefIt, it achieves \textbf{77.6\%} precision at IoU 0.5, a numerical margin of \textbf{5.2 percentage points} over the reported state of the art under differing evaluation protocols. Ablations support joint geometric and visual supervision on the standard and similar-object sets, and favor parity over three alternative pooling operators on the standard set. These results support task-structured supervision for continuous pointing grounding. We will release the code and supporting materials.
\end{abstract}

\begingroup
\maintextspacing

\section{Introduction}
\label{sec:intro}
\emph{Embodied reference understanding} identifies the object a person refers to through language and gesture in a shared environment~\citep{chen2021yourefit}. We focus on its image-based pointing setting: given an image and a language query, the model predicts the referred object's bounding box~\citep{li2026egopoint}. Consider several similar cups on a table and the query ``this cup.'' Recognizing the cups does not resolve which instance is intended; interpreting the hand's direction does not determine the object's full extent. Resolving the reference requires connecting gesture geometry with visual evidence of the target's identity and spatial extent.

\begin{figure}[t]
\centering
\includegraphics[width=\linewidth]{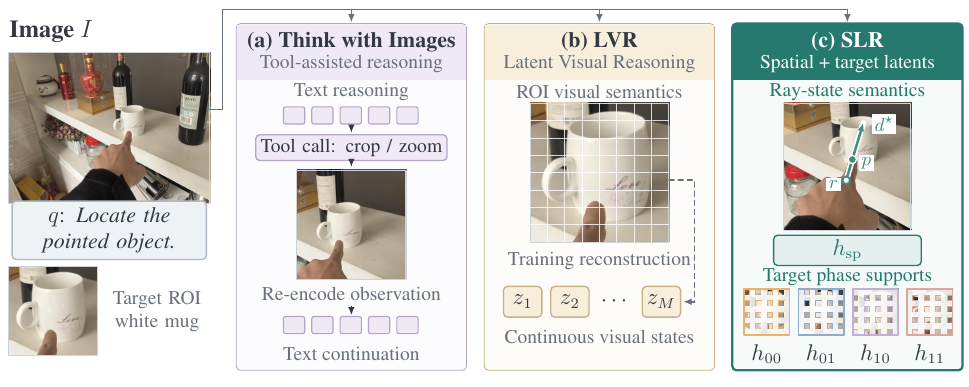}
\caption{Comparison of reasoning approaches on an EgoPoint-Ground example~\citep{li2026egopoint}. (a) Tool-assisted reasoning crops and re-encodes image regions during text generation~\citep{zheng2025deepeyes}. (b) LVR~\citep{li2025lvr} supervises continuous states by reconstructing annotated region features. (c) SLR combines one spatial ray state with four states supervised by parity-pooled target features. Image crops and ray overlays illustrate the reasoning operations and supervision targets; generic latent boundary markers are omitted for clarity.}
\label{fig:comparison}
\end{figure}

Intermediate reasoning offers a way to organize these cues before predicting the intended referent. Explicit approaches generate textual steps or invoke visual tools~\citep{hu2024sketchpad,zheng2025deepeyes}; continuous approaches feed generated hidden states back as input embeddings~\citep{hao2024coconut}. LVR~\citep{li2025lvr} supervises visual-feature reconstruction with teacher-forced region tokens, while RIS~\citep{cui2026ris} combines box and semantic supervision with staged internalization. These methods establish region-grounded supervision as a foundation for latent reasoning. Embodied reference understanding with pointing gestures requires modeling a specific relation: how the observed hand indicates the intended referent. Region reconstruction and box prediction do not directly supervise that hand geometry. A fixed latent budget also requires a choice of how to distribute target-region evidence across states. This motivates our central question: \emph{how should gesture geometry and target appearance supervise latent reasoning for embodied reference understanding?}

In this paper, we propose \emph{Spatial Latent Reasoning (SLR)}, organizing gesture geometry and target appearance into ordered, model-generated continuous states. A \emph{spatial ray state} is trained through a lightweight auxiliary head to predict the fingertip as the ray origin and the finger-base-to-fingertip direction as its orientation. Direction is defined on the visual-token grid to account for nonsquare layouts. These labels depend only on the observed hand, providing supervision distinct from target-region reconstruction.

The recurrence carries the full-dimensional spatial state; the head's predictions are not fed back. Geometric supervision, subsequent visual alignment, and answer generation jointly train this state to retain task-relevant context alongside pointing information. Four subsequent \emph{target visual states} align with referred-region features, conditioning their generation on the geometry-supervised state. Training and inference share this model-generated recurrence; auxiliary labels enter only the losses.

We introduce \emph{parity pooling} to distribute target-region evidence across four visual states. Drawing on region aggregation~\citep{he2014spp} and polyphase decomposition~\citep{smith2011spectral}, it groups tokens by global row and column parity and averages each group. The resulting targets summarize interleaved positions across a variable-size ROI, providing distributed regional evidence without crop re-encoding or textual descriptions. Comparisons with quadrant grouping, random grouping, and max aggregation evaluate this construction. Figure~\ref{fig:comparison} contrasts the reasoning paths; Appendix~\ref{app:related} provides further comparisons.

Our contributions are threefold: (1) We propose \textbf{SLR}, an ordered supervision scheme that couples a full-dimensional, geometry-supervised ray state with subsequent target visual states for embodied reference understanding. (2) We introduce \textbf{parity pooling} for latent supervision, converting variable-size target regions into four phase-specific feature targets with interleaved spatial support. (3) SLR improves EgoPoint-Ground mIoU over same-backbone SFT by 2.8, 17.5, and 21.1 percentage points on Qwen3.5-4B, Qwen2.5-VL-7B, and Qwen3-VL-8B, respectively, with gains on both hard subsets. On YouRefIt, it reaches 77.6\% precision at IoU 0.5.

\section{Method}
\label{sec:method}

Building on Latent Visual Reasoning (LVR)~\citep{li2025lvr}, SLR generates one spatial ray state and four target visual states before decoding the bounding box:
\begin{equation}
 (I,q)\longrightarrow h_{\mathrm{sp}}
 \longrightarrow h_{00}\longrightarrow h_{01}
 \longrightarrow h_{10}\longrightarrow h_{11}
 \longrightarrow y_{\mathrm{bbox}}.
 \label{eq:reasoning_sequence}
\end{equation}
Figure~\ref{fig:architecture} summarizes the geometric and visual supervision of this sequence.

\subsection{Problem Formulation}
\label{sec:problem}
Let $I\in\mathbb{R}^{H_0\times W_0\times 3}$ be an image with a pointing gesture and $q$ a grounding query. We learn
\begin{equation}
 f_\theta:(I,q)\mapsto \widehat b, \qquad
 b=(x_1,y_1,x_2,y_2)\in\mathcal{B},
\end{equation}
where $\mathcal{B}=\{b\in[0,1]^4:x_1<x_2,\;y_1<y_2\}$ is the space of normalized, axis-aligned boxes. Training examples form $\mathcal{D}=\{(I_n,q_n,b_n^\star)\}_{n=1}^{N}$, where $b_n^\star$ is the annotated target box. We serialize each box as an answer sequence $y^\star=\tau(b^\star)$ using integer coordinates on a $1000\times1000$ grid.

Training examples may also provide finger-base and fingertip annotations $(r_n,p_n)$. Geometry and target-region features supply auxiliary losses only and are never inputs to the generated states.

\paragraph{Model representations.}
A frozen visual encoder and projector $E_\phi$ map the image to a post-merge grid $V=E_\phi(I)\in\mathbb{R}^{H_g\times W_g\times d}$. The multimodal prompt $P(I,q)$ packs these visual tokens with query embeddings and modality/position markers. A causal language backbone $F_\theta$ generates continuous states $Z_\theta(I,q)=(z_1,\ldots,z_K)$, where $z_k\in\mathbb{R}^d$, before the box answer. Appendix~\ref{app:latent_background} gives the autoregressive formulation and the original LVR supervision objective.

\subsection{Recurrent Continuous Latent States}
\label{sec:recurrence}
Given the multimodal prompt $P(I,q)$, a latent-start marker $s$, and a latent-end marker $e$, each continuous slot receives the final hidden vector from the preceding position:
\begin{equation}
 \begin{aligned}
 z_1 &= F_\theta\!\left([P(I,q),s]\right)_{\mathrm{last}},\\
 z_k &= F_\theta\!\left([P(I,q),s,z_1,\ldots,z_{k-1}]\right)_{\mathrm{last}},
 \quad 2\leq k\leq K.
 \end{aligned}
 \label{eq:recurrence}
\end{equation}
Here $F_\theta(\cdot)_{\mathrm{last}}$ is the final-layer hidden vector before vocabulary projection. We use $K=5$, with
\begin{equation}
 (z_1,z_2,z_3,z_4,z_5)
 =(h_{\mathrm{sp}},h_{00},h_{01},h_{10},h_{11}),
 \label{eq:slot_roles}
\end{equation}
and decode the answer after $[P,s,z_1,z_2,z_3,z_4,z_5,e]$. All states retain dimension $d$ ($2560$ for Qwen3.5-4B). Slot roles follow their order and supervision; we retain Coconut's generic boundary markers without role-specific markers or embeddings.

Auxiliary losses supervise the producer vectors $z_k$ injected into the slots, rather than the outputs at those slots after recomputation. Gradients propagate through the recurrence; causal masking prevents later placeholders and teacher-forced answer tokens from influencing earlier states.

\begin{figure}[t]
\centering
\includegraphics[width=\linewidth]{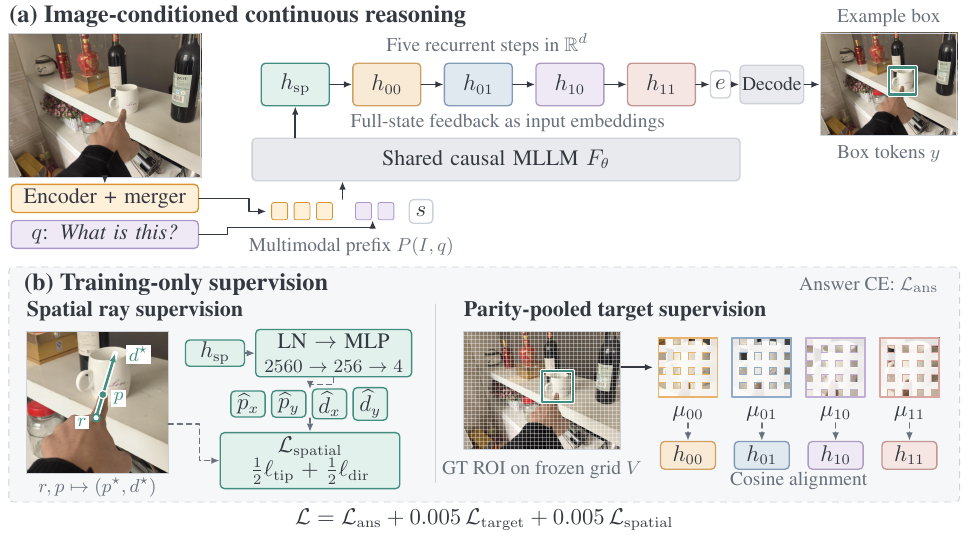}
\caption{Overview of SLR. (a) A frozen visual encoder and merger provide image features. The causal MLLM generates five recurrent states in its hidden space, then decodes the target box; $s$ and $e$ denote latent boundary markers. (b) A spatial readout predicts fingertip position and pointing direction, while visual states align with parity-pooled region features with stopped gradients. Auxiliary readouts and annotations are used only for training. The ray overlays and grid are schematic; readout dimensions correspond to Qwen3.5-4B.}
\label{fig:architecture}
\end{figure}

\subsection{Spatial Ray-State Latent}
\label{sec:spatial_semantic}
An auxiliary head supervises the origin and orientation of a pointing ray from $h_{\mathrm{sp}}$. Position anchors the gesture in the image, while direction specifies its orientation; together they describe a pointing cue without selecting a candidate object. The full-dimensional state also receives gradients from the subsequent visual and answer objectives, allowing the recurrence to combine this geometric cue with evidence needed for referent selection.

\paragraph{Geometric supervision in two coordinate systems.}
Let $r=(x_r,y_r)$ and $p=(x_p,y_p)$ denote the annotated finger base and fingertip in normalized-1000 image coordinates, with the horizontal axis pointing right and the vertical axis pointing down. The fingertip regression target lies in $[-1,1]^2$:
\begin{equation}
 p^\star=\left(2\frac{x_p}{1000}-1,\;2\frac{y_p}{1000}-1\right).
 \label{eq:tip_target}
\end{equation}
We scale the finger-base-to-fingertip displacement to the post-merge grid before normalizing:
\begin{equation}
 \Delta_g=\left(W_g\frac{x_p-x_r}{1000},\;H_g\frac{y_p-y_r}{1000}\right),
 \qquad d^\star=\frac{\Delta_g}{\max(\|\Delta_g\|_2,\epsilon)},
 \qquad \epsilon=10^{-12}.
 \label{eq:spatial_targets}
\end{equation}
This scaling accounts for nonsquare grids; normalizing afterward removes displacement magnitude while preserving orientation in the visual grid. The position and direction targets use different scales; expressing both in grid coordinates defines the pointing ray
\begin{equation}
 p_g=\left(W_g\frac{x_p}{1000},\;H_g\frac{y_p}{1000}\right),
 \qquad \mathcal{R}(t)=p_g+t\,d^\star,\quad t\geq0.
 \label{eq:spatial_ray}
\end{equation}
The finger base determines direction only; the spatial labels do not use target boxes or candidate objects. Answer and target-alignment gradients can nevertheless make the state target-aware.

\paragraph{Auxiliary readout.}
A lightweight training head reads $h_{\mathrm{sp}}$ using LayerNorm and two linear maps:
\begin{equation}
 \begin{aligned}
 o&=W_2\,\operatorname{GELU}\!\left(W_1\,\operatorname{LN}(h_{\mathrm{sp}})+b_1\right)+b_2,\\
 \widehat p&=o_{1:2},\qquad
 \widehat d=\frac{o_{3:4}}{\max(\|o_{3:4}\|_2,\epsilon)}.
 \end{aligned}
 \label{eq:spatial_head}
\end{equation}
The head maps $d\rightarrow256\rightarrow4$, with unconstrained fingertip outputs and a normalized direction. Its predictions are not injected into the recurrent path.

\paragraph{Spatial loss and validity masking.}
For each valid example, we average SmoothL1 over the two fingertip coordinates and combine it with directional cosine distance:
\begin{equation}
 \ell_{\mathrm{spatial}}
 =\tfrac12\operatorname{SmoothL1}_{\beta=0.1}(\widehat p,p^\star)
 +\tfrac12(1-\widehat d^\top d^\star).
 \label{eq:spatial_sample_loss}
\end{equation}
Let $\mathcal{B}_{\mathrm{sp}}$ contain minibatch examples with finite finger-base and fingertip annotations and $\|p-r\|_2>10^{-6}$ in normalized-1000 coordinates. Averaging $\ell_{\mathrm{spatial}}$ over this set gives $\mathcal{L}_{\mathrm{spatial}}$; an empty set gives zero loss. Examples with invalid finger geometry skip only the spatial auxiliary term; their answer loss and any valid target-region supervision remain active.

\subsection{Parity-Pooled Visual Targets}
\label{sec:parity}

\begin{figure}[t]
\centering
\includegraphics[width=\linewidth]{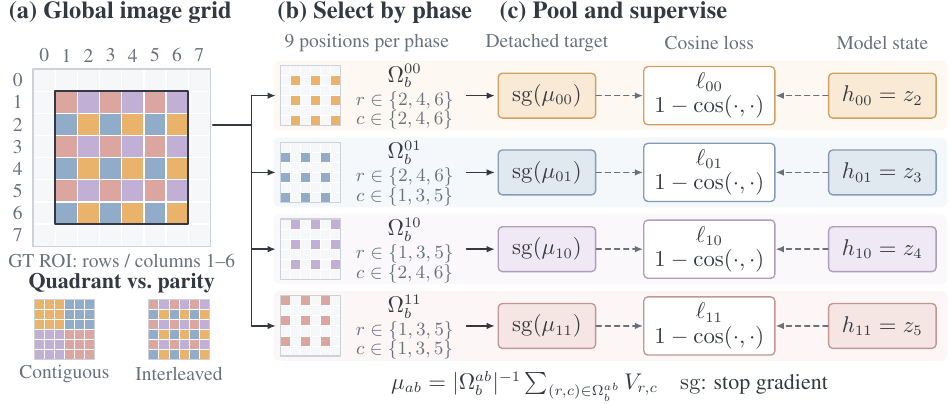}
\caption{Global-grid parity pooling. The target ROI spans rows and columns 1--6 of an illustrative $8\times8$ grid. Each branch averages nine feature vectors selected by global row and column parity and aligns the detached mean with its generated state $h_{ab}=z_{2+2a+b}$. Dashed arrows connect both inputs to the cosine loss; pooled features serve only as training targets. The inset compares contiguous quadrants with interleaved parity groups.}
\label{fig:parity}
\end{figure}
\paragraph{Target support.}
On the post-merge visual grid $V$, token $(r,c)$ has normalized center $u_{r,c}=((c+\tfrac12)/W_g,(r+\tfrac12)/H_g)$. We select tokens whose centers lie inside the target box:
\begin{equation}
 \Omega_b=\{(r,c):u_{r,c}\in[x_1,x_2]\times[y_1,y_2]\},
 \qquad M_b(r,c)=\mathbf{1}[(r,c)\in\Omega_b].
 \label{eq:roi}
\end{equation}
The support is defined directly by the annotated bounding box.

\paragraph{Polyphase decomposition.}
In multirate signal processing, a discrete signal is decomposed into subsequences indexed by sampling phase~\citep{smith2011spectral}. For a two-dimensional signal $X$, stride-two sampling along each axis gives four polyphase components:
\begin{equation}
 (\mathcal{P}_{\alpha\beta}X)_{i,j}=X_{2i+\alpha,\,2j+\beta},
 \qquad (\alpha,\beta)\in\{0,1\}^{2}.
 \label{eq:polyphase}
\end{equation}
The components partition the lattice, as in polyphase representations for image sampling~\citep{chaman2021shift}; we partition the ROI mask and visual features to define supervision targets.

\paragraph{Region-conditioned phase averages.}
For phase $(\alpha,\beta)$, define
\begin{equation}
 \Omega_b^{\alpha\beta}=\{(r,c)\in\Omega_b:r\bmod2=\alpha,\;c\bmod2=\beta\},
 \qquad n_{\alpha\beta}=|\Omega_b^{\alpha\beta}|.
 \label{eq:phase_support}
\end{equation}
For each nonempty support ($n_{\alpha\beta}>0$), parity pooling computes a mean feature
\begin{equation}
 \begin{aligned}
 \mu_{\alpha\beta}(V,b)
 &=\frac{1}{n_{\alpha\beta}}\sum_{(r,c)\in\Omega_b^{\alpha\beta}} V_{r,c}\\
 &=\frac{\sum_{i,j}(\mathcal{P}_{\alpha\beta}M_b)_{i,j}
                    (\mathcal{P}_{\alpha\beta}V)_{i,j}}
          {\sum_{i,j}(\mathcal{P}_{\alpha\beta}M_b)_{i,j}}.
 \end{aligned}
 \label{eq:parity_pool}
\end{equation}
The descriptors supervise $(z_2,z_3,z_4,z_5)$ in order $(00,01,10,11)$. The four phase indices are anchored to the full image grid and remain fixed when the target ROI changes.

Figure~\ref{fig:parity} contrasts these interleaved supports with contiguous quadrants. For example, phase $(0,1)$ pools tokens at even rows and odd columns throughout the ROI, rather than selecting one side of the object. The fixed phase-to-state correspondence assigns each recurrent state a consistent sampling rule across differently sized regions.

Each ROI token contributes to exactly one phase. Empty phases are omitted from the auxiliary loss without duplicating tokens, while all four recurrent slots remain. The phase means recover the overall ROI mean when weighted by their counts, but discard within-group arrangement. Appendix~\ref{app:parity_properties} gives the occupancy and mean-recovery identities.

\subsection{Visual Alignment and Joint Training}
\label{sec:objective}
Let $\mu_{n,k}$ denote the phase mean assigned to state $z_{n,k+1}$, and let $\mathcal{A}$ collect minibatch example--slot pairs with nonempty phase supports. Using stop-gradient targets, we minimize
\begin{equation}
\mathcal{L}_{\mathrm{target}}=
 \frac{1}{|\mathcal{A}|}\sum_{(n,k)\in\mathcal{A}}
 \left(1-\frac{z_{n,k+1}^{\top}\operatorname{sg}(\mu_{n,k})}
 {\|z_{n,k+1}\|_2\,\|\operatorname{sg}(\mu_{n,k})\|_2}\right),
 \label{eq:alignment}
\end{equation}
with stabilized norms and zero loss if $\mathcal{A}$ is empty. The average is over active pairs, and cosine alignment constrains direction rather than magnitude.

Let $\mathcal{T}$ be the set of supervised answer-token positions in the minibatch. The complete objective is
\begin{equation}
 \begin{aligned}
 \mathcal{L}_{\mathrm{ans}}
 &=-\frac{1}{|\mathcal{T}|}\sum_{(n,t)\in\mathcal{T}}
 \log p_\theta(y_{n,t}^{\star}\mid I_n,q_n,Z_\theta(I_n,q_n),y_{n,<t}^{\star}),\\
 \mathcal{L}&=\mathcal{L}_{\mathrm{ans}}+\lambda_{\mathrm{target}}\mathcal{L}_{\mathrm{target}}
 +\lambda_{\mathrm{spatial}}\mathcal{L}_{\mathrm{spatial}},
 \qquad \lambda_{\mathrm{target}}=\lambda_{\mathrm{spatial}}=0.005.
 \end{aligned}
 \label{eq:objective}
\end{equation}
Only the serialized box and end-of-sequence token contribute teacher-forced cross-entropy; all prompt, marker, latent, and padding positions are masked. We freeze the visual encoder and merger/projector, and train the language-side parameters, including text embeddings, and spatial head. All five recurrent states are model-generated. Answer and target-alignment gradients propagate through earlier states, coupling intermediate supervision to the final grounding objective.

\subsection{Inference}
\label{sec:inference}
Given $(I,q)$, the model generates five states, appends the latent-end marker, and greedily decodes the box until the end-of-sequence token or output limit. The spatial readout is unused, and no ray--object intersection is performed. Compute depends on prompt and answer length, caching, and the fixed latent budget. Appendices~\ref{app:decoding} and~\ref{app:metrics} detail parsing, constrained decoding, and timing.

\section{Experiments}
\label{sec:experiments}

\begin{table*}[!htb]
\centering
\caption{Results on \textbf{EgoPoint-Ground}. SLR uses one spatial and four parity-supervised visual states. Shading marks SLR; bold denotes the best grounding scores per backbone, including ties. Human scores provide a reference; token counts and times appear in Table~\ref{tab:computation_records} in Appendix~\ref{app:computation_records}.}
\label{tab:main_egopoint}
\papertablesetup
\begin{tabular*}{\textwidth}{@{\extracolsep{\fill}}llrrrr@{}}
\toprule
Model & Method & P@0.3 $\uparrow$ & P@0.5 $\uparrow$ & P@0.7 $\uparrow$ & mIoU $\uparrow$ \\
\midrule
Human Study & Human Participants & 0.922 & 0.909 & 0.830 & 0.805 \\
\midrule
\multirow{5}{*}{Qwen2.5-VL-7B} & Zero-shot & 0.650 & 0.589 & 0.500 & 0.546 \\
& Standard SFT & 0.644 & 0.589 & 0.489 & 0.538 \\
& Text CoT~\citep{wei2022cot} (adapted) & 0.744 & 0.661 & 0.494 & 0.575 \\
& PointVG-R~\citep{li2026pointvgr} & 0.833 & 0.761 & 0.639 & 0.670 \\
\rowcolor{oursrow}
\cellcolor{white}& \textbf{SLR} & \textbf{0.839} & \textbf{0.806} & \textbf{0.706} & \textbf{0.713} \\
\midrule
\multirow{4}{*}{Qwen3-VL-8B} & Zero-shot & 0.556 & 0.528 & 0.506 & 0.516 \\
& Standard SFT & 0.594 & 0.550 & 0.517 & 0.533 \\
& Text CoT~\citep{wei2022cot} (adapted) & 0.794 & 0.778 & 0.717 & 0.701 \\
\rowcolor{oursrow}
\cellcolor{white}& \textbf{SLR} & \textbf{0.828} & \textbf{0.800} & \textbf{0.761} & \textbf{0.744} \\
\midrule
\multirow{5}{*}{Qwen3.5-4B} & Zero-shot (Non-thinking) & 0.356 & 0.322 & 0.289 & 0.332 \\
& Zero-shot (Thinking) & 0.578 & 0.544 & 0.456 & 0.484 \\
& Standard SFT & 0.861 & 0.839 & 0.761 & 0.750 \\
& LVR~\citep{li2025lvr} & 0.600 & 0.561 & 0.511 & 0.525 \\
\rowcolor{oursrow}
\cellcolor{white}& \textbf{SLR} & \textbf{0.906} & \textbf{0.878} & \textbf{0.811} & \textbf{0.778} \\
\bottomrule
\end{tabular*}
\end{table*}

\subsection{Datasets and Evaluation}

We evaluate on \textbf{EgoPoint-Ground}~\citep{li2026egopoint}, which provides egocentric hand--target annotations, and \textbf{YouRefIt}~\citep{chen2021yourefit}, an embodied reference benchmark with language and gesture. Within EgoPoint-Ground, \emph{Hard-Similar} focuses on visually similar objects and \emph{Hard-Complex} on complex scenes, allowing evaluation under two challenging grounding conditions.

We report mIoU and localization accuracy P@$\tau$ on EgoPoint-Ground at $\tau\in\{0.3,0.5,0.7\}$. YouRefIt uses thresholds $\{0.25,0.5,0.75\}$; each row identifies its evaluation metric. We organize the main comparisons by backbone and use mIoU to summarize box overlap across EgoPoint-Ground examples. Thresholded accuracy additionally measures localization at increasing precision requirements. Table~\ref{tab:computation_records} reports token counts and times; Appendix~\ref{app:metrics} details their measurement.

\subsection{Comparison Methods}
The comparison covers direct prediction (zero-shot and standard SFT), explicit reasoning (Text CoT~\citep{wei2022cot} and PointVG-R~\citep{li2026pointvgr}), and continuous visual reasoning (LVR~\citep{li2025lvr}). Tables~\ref{tab:main_egopoint} and~\ref{tab:main_hard} show the available comparisons on Qwen2.5-VL-7B, Qwen3-VL-8B, and Qwen3.5-4B. Rows marked ``adapted'' denote backbone-specific adaptations. Tables group methods by backbone to distinguish improvements from the proposed framework from differences in the underlying model; human performance is reported separately as a reference.

\subsection{Implementation and Checkpoint Selection}
\label{sec:implementation}
Our Qwen3.5-4B implementation freezes the visual encoder and merger and trains the language model and spatial readout. It uses AdamW with learning rate $10^{-5}$, weight decay $0.01$, and an effective batch size of 12 across four devices. Table~\ref{tab:reference_config} summarizes the reference settings; Appendix~\ref{app:implementation} details the architecture, training, and preprocessing.

We select checkpoints by validation mIoU using free greedy generation with a 96-token output limit and caching disabled. The highest-scoring checkpoint is selected, with the earlier epoch breaking an exact tie, and is fixed across the parent test set and both hard subsets. This protocol separates checkpoint selection from final evaluation and avoids subset-specific checkpoint selection.

\begin{table*}[!htb]
\centering
\caption{Key comparisons on the two \textbf{EgoPoint-Ground} hard subsets. Shading identifies SLR; bold scores mark the best reported result within each backbone and subset, including ties. Table~\ref{tab:hard_full} in Appendix~\ref{app:hard_results} reports all methods, localization metrics, token counts, and times.}
\label{tab:main_hard}
\papertablesetup
\begin{tabular*}{\textwidth}{@{\extracolsep{\fill}}llrrrr@{}}
\toprule
\multirow{2}{*}{Model} & \multirow{2}{*}{Method} & \multicolumn{2}{c}{Hard-Similar} & \multicolumn{2}{c}{Hard-Complex} \\
\cmidrule(lr){3-4}\cmidrule(l){5-6}
 & & P@0.5 $\uparrow$ & mIoU $\uparrow$ & P@0.5 $\uparrow$ & mIoU $\uparrow$ \\
\midrule
\multirow{4}{*}{Qwen2.5-VL-7B} & Standard SFT & 0.229 & 0.202 & 0.458 & 0.389 \\
 & Text CoT~\citep{wei2022cot} (adapted) & 0.188 & 0.198 & 0.448 & 0.411 \\
 & PointVG-R~\citep{li2026pointvgr} & 0.313 & 0.276 & 0.573 & 0.473 \\
\rowcolor{oursrow}
\cellcolor{white} & \textbf{SLR} & \textbf{0.323} & \textbf{0.327} & \textbf{0.583} & \textbf{0.511} \\
\midrule
\multirow{3}{*}{Qwen3-VL-8B} & Standard SFT & 0.281 & 0.290 & 0.385 & 0.349 \\
 & Text CoT~\citep{wei2022cot} (adapted) & \textbf{0.333} & \textbf{0.353} & 0.500 & 0.452 \\
\rowcolor{oursrow}
\cellcolor{white} & \textbf{SLR} & 0.302 & 0.315 & \textbf{0.542} & \textbf{0.495} \\
\midrule
\multirow{3}{*}{Qwen3.5-4B} & Standard SFT & 0.396 & 0.361 & 0.594 & 0.536 \\
 & LVR~\citep{li2025lvr} & 0.167 & 0.158 & 0.396 & 0.396 \\
\rowcolor{oursrow}
\cellcolor{white} & \textbf{SLR} & \textbf{0.458} & \textbf{0.421} & \textbf{0.656} & \textbf{0.551} \\
\bottomrule
\end{tabular*}
\end{table*}

\subsection{Main Results}
\paragraph{Improvements across models and localization criteria.}
Tables~\ref{tab:main_egopoint} and~\ref{tab:main_hard} show higher mIoU than same-backbone SFT in all nine backbone--evaluation combinations. Standard-set gains range from 2.8 to 21.1 percentage points, supporting the framework across different backbones. The benefit also extends beyond the direct-prediction baseline: on Qwen2.5-VL-7B and Qwen3-VL-8B, SLR exceeds the strongest reported reasoning baseline in standard-set mIoU by 4.3 points in each case. SLR leads all three localization thresholds within each backbone on the standard set. For Qwen3.5-4B, P@0.7 rises from 0.761 to 0.811, showing that the gain includes boxes meeting a stricter overlap criterion. This supports improved localization quality, although aggregate IoU metrics do not separate referent-selection errors from box-placement errors.

\paragraph{Performance depends on the grounding condition.}
Both hard subsets retain mIoU gains over SFT across all three backbones. Qwen3.5-4B gains more on Hard-Similar than Hard-Complex (6.0 versus 1.5 points), making visually ambiguous referents a particularly informative setting for the component analysis below. The two subsets also expose differences between metrics. On Qwen2.5-VL-7B Hard-Complex, PointVG-R leads at P@0.3 (0.656 versus 0.615), while SLR leads at P@0.7 (0.469 versus 0.365) and mIoU (0.511 versus 0.473). Thus, the ranking changes as the overlap requirement becomes stricter. On Qwen3-VL-8B Hard-Similar, Text CoT has higher mIoU (0.353 versus 0.315), indicating that the advantage over explicit reasoning is setting-dependent. Table~\ref{tab:hard_full} in Appendix~\ref{app:hard_results} gives the full results. Figures~\ref{fig:qualitative_comparison} and~\ref{fig:latent_attention} illustrate predictions and attention.

\paragraph{Evidence on a second benchmark.}
On YouRefIt, the gains over the same-backbone zero-shot model are similar across IoU thresholds 0.25/0.5/0.75: 8.4/8.0/8.1 points (Table~\ref{tab:main_yourefit}). The persistence of the gain at the strictest threshold supports an improvement in precise grounding on this benchmark. This comparison measures the trained framework against the pretrained model; it does not isolate the contribution beyond task-specific SFT. Literature scores retain their original protocols.

\begin{table*}[!htb]
\centering
\caption{Results on \textbf{YouRefIt}. Published baselines use their original metrics and evaluation protocols. Shading identifies SLR; the zero-shot model uses the same backbone.}
\label{tab:main_yourefit}
\papertablesetup
\begin{tabular*}{\textwidth}{@{\extracolsep{\fill}}lcrrr@{}}
\toprule
Method & Reported Metric & @0.25 $\uparrow$ & @0.50 $\uparrow$ & @0.75 $\uparrow$ \\
\midrule
Human Participants~\citep{chen2021yourefit} & Accuracy & 0.942 & 0.858 & 0.533 \\
\midrule
YouRefIt Full~\citep{chen2021yourefit} & Accuracy & 0.547 & 0.405 & 0.140 \\
REP~\citep{shi2022rep} & Precision & 0.588 & 0.457 & 0.188 \\
Touch-Line (VTL)~\citep{li2023touchline} & Precision & 0.711 & 0.635 & 0.390 \\
AD-DINO (ADTL)~\citep{guo2024addino} & Accuracy & 0.763 & 0.724 & 0.554 \\
CAPE~\citep{eyiokur2026cape} & mAP & 0.750 & 0.654 & 0.357 \\
DA-ERU~\citep{eyiokur2026daeru} & mAP & 0.787 & 0.676 & 0.381 \\
\midrule
Zero-shot (Qwen3.5-4B) & Precision & 0.766 & 0.696 & 0.460 \\
\rowcolor{oursrow}
\textbf{SLR (Qwen3.5-4B)} & Precision & 0.850 & 0.776 & 0.541 \\
\bottomrule
\end{tabular*}
\end{table*}

\begin{figure}[!htb]
\centering
\includegraphics[width=\linewidth]{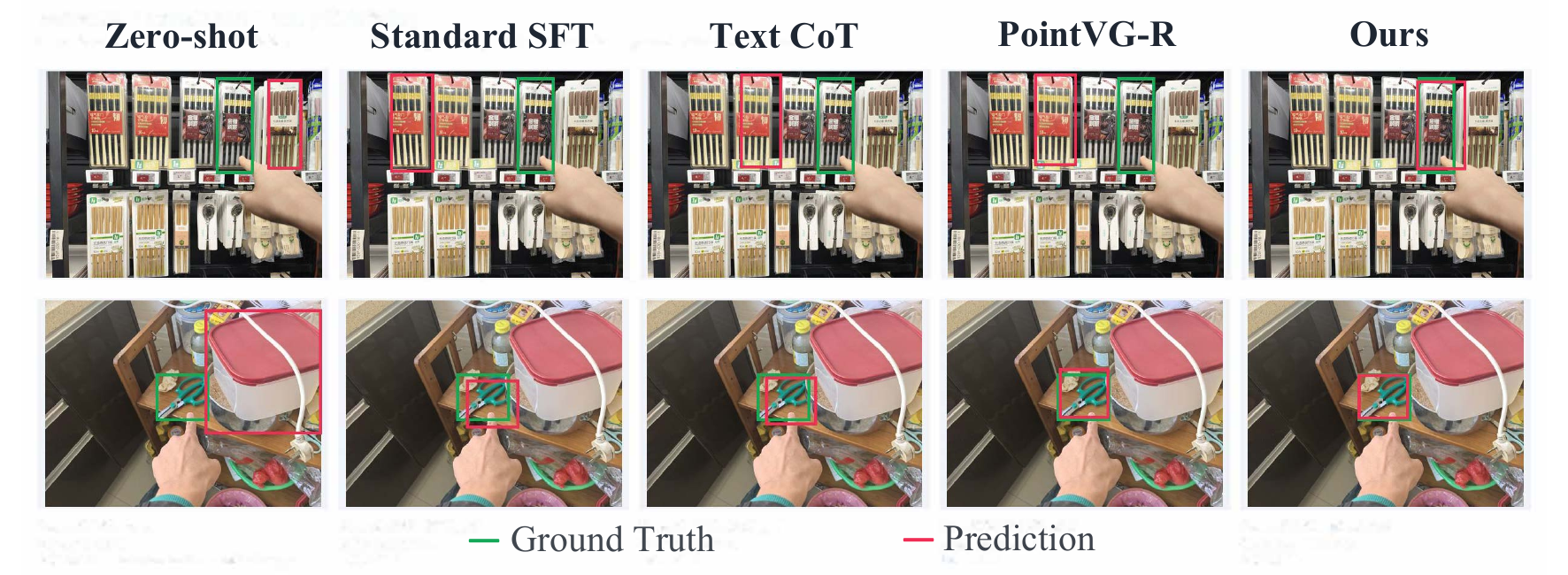}
\caption{Grounding comparisons for two pointing examples. Columns show Zero-shot, Standard SFT, Text CoT, PointVG-R, and SLR (labeled ``Ours'' in the supplied visualization). Green and red boxes indicate the annotated targets and predicted regions, respectively, for each of the five compared methods.}
\label{fig:qualitative_comparison}
\end{figure}

\paragraph{Qualitative grounding.}
Figure~\ref{fig:qualitative_comparison} illustrates referent selection among similar products and localization in a cluttered scene. Our predictions overlap the annotated targets in both examples, complementing the aggregate results with concrete instances of spatial grounding.

\subsection{Ablation Studies}
Using Qwen3.5-4B, we evaluate components across all three EgoPoint-Ground sets and pooling on the standard set. Figure~\ref{fig:latent_budget} and Table~\ref{tab:ablation_budget} in Appendix~\ref{app:budget_plot} report the target-state count sweep.

\begin{table*}[!htb]
\centering
\caption{Component ablations on \textbf{Qwen3.5-4B}. Shading marks the full model (one spatial and four target states); bold marks the best grounding scores per subset, including ties. Spatial-only retains the ray state; Target-only retains the four visual states; SFT uses no latent segment.}
\label{tab:ablation_components}
\papertablesetup
\begin{tabular*}{\textwidth}{@{\extracolsep{\fill}}llrrrr@{}}
\toprule
Backbone & Variant & P@0.3 $\uparrow$ & P@0.5 $\uparrow$ & P@0.7 $\uparrow$ & mIoU $\uparrow$ \\
\midrule
\multicolumn{6}{@{}l}{\textbf{EgoPoint-Ground}} \\
\midrule
\rowcolor{oursrow}
\cellcolor{white}\multirow{4}{*}{Qwen3.5-4B} & \textbf{Spatial + Parity (SLR)} & \textbf{0.906} & \textbf{0.878} & \textbf{0.811} & \textbf{0.778} \\
 & Spatial-only & 0.867 & 0.844 & 0.767 & 0.747 \\
 & Target-only & 0.883 & 0.856 & 0.783 & 0.764 \\
 & SFT (no latent states) & 0.861 & 0.839 & 0.761 & 0.750 \\
\midrule
\multicolumn{6}{@{}l}{\textbf{Hard-Similar: multiple visually similar objects}} \\
\midrule
\rowcolor{oursrow}
\cellcolor{white}\multirow{4}{*}{Qwen3.5-4B} & \textbf{Spatial + Parity (SLR)} & \textbf{0.500} & \textbf{0.458} & \textbf{0.396} & \textbf{0.421} \\
 & Spatial-only & 0.323 & 0.302 & 0.250 & 0.286 \\
 & Target-only & 0.427 & 0.396 & 0.344 & 0.369 \\
 & SFT (no latent states) & 0.417 & 0.396 & 0.344 & 0.361 \\
\midrule
\multicolumn{6}{@{}l}{\textbf{Hard-Complex: complex scenes}} \\
\midrule
\rowcolor{oursrow}
\cellcolor{white}\multirow{4}{*}{Qwen3.5-4B} & \textbf{Spatial + Parity (SLR)} & 0.719 & 0.656 & 0.479 & 0.551 \\
 & Spatial-only & 0.688 & 0.625 & 0.479 & 0.558 \\
 & Target-only & \textbf{0.760} & \textbf{0.677} & \textbf{0.521} & \textbf{0.597} \\
 & SFT (no latent states) & 0.667 & 0.594 & 0.458 & 0.536 \\
\bottomrule
\end{tabular*}
\end{table*}

\paragraph{Geometry is most effective with visual supervision on Hard-Similar.}
Table~\ref{tab:ablation_components} shows that the full model achieves the best mIoU on the standard and Hard-Similar sets. Adding the spatial state to Target-only yields a larger gain on Hard-Similar than on the standard set (5.2 versus 1.4 points). Crucially, Spatial-only falls below SFT on Hard-Similar (0.286 versus 0.361), whereas the joint model reaches 0.421. The strongest result therefore comes from supervising both cues together, consistent with the motivation that pointing direction and target appearance provide complementary information when similar objects compete as referents. The Hard-Similar gain also holds at the strictest overlap threshold: joint supervision raises P@0.7 from 0.344 for Target-only to 0.396. Thus, the improvement extends to more precisely localized predictions. Spatial-only does not show this gain, suggesting that a directional cue is most useful when paired with supervision of the target's visual extent. On Hard-Complex, Target-only leads the full model (0.597 versus 0.551), showing that this benefit depends on scene conditions. These configuration comparisons do not isolate the causal contribution of a specific geometric mechanism.

\begin{table*}[!htb]
\centering
\caption{Pooling-operator ablations on \textbf{EgoPoint-Ground} with Qwen3.5-4B, four target latents, and no spatial state. Shading identifies parity pooling; bold scores mark the best grounding result, including ties. Appendix~\ref{app:pooling_controls} defines the operators.}
\label{tab:ablation_pooling}
\papertablesetup
\begin{tabular*}{\textwidth}{@{\extracolsep{\fill}}llrrrr@{}}
\toprule
Backbone & Pooling & P@0.3 $\uparrow$ & P@0.5 $\uparrow$ & P@0.7 $\uparrow$ & mIoU $\uparrow$ \\
\midrule
\rowcolor{oursrow}
\cellcolor{white}\multirow{4}{*}{Qwen3.5-4B} & \textbf{Parity} & \textbf{0.883} & \textbf{0.856} & \textbf{0.783} & \textbf{0.764} \\
& Quadrant        & 0.839 & 0.817 & 0.744 & 0.728 \\
& Random          & 0.844 & 0.822 & 0.756 & 0.735 \\
& Max             & 0.850 & 0.833 & 0.756 & 0.746 \\
\bottomrule
\end{tabular*}
\end{table*}

\paragraph{Target construction matters at a fixed state budget.}
Table~\ref{tab:ablation_pooling} compares four target states without a spatial state. Parity exceeds Quadrant, Random, and Max in mIoU by 3.6, 2.9, and 1.8 points and leads at all three thresholds. Since all variants use four states, the advantage is not explained by a larger latent-step budget. The controls probe different aspects of the target construction: Quadrant changes interleaved support to contiguous regions, Random changes group membership, and Max changes aggregation on the parity supports (Appendix~\ref{app:pooling_controls}). Their ordering favors the combination of spatial interleaving and mean aggregation on the standard set. Each parity target aggregates interleaved locations throughout the ROI, preserving phase-specific differences while providing distributed regional evidence. Mean aggregation incorporates all sampled features, whereas Max retains their componentwise extrema. The budget sweep (Appendix~\ref{app:budget_plot}) gives a complementary result: four states lead in mIoU, but one state is close (0.764 versus 0.761), and larger budgets bring no improvement. Together, these findings support a compact supervision budget with deliberately constructed visual targets and identify target organization as an important design choice for recurrent visual grounding.

\begin{figure}[!htb]
\centering
\includegraphics[width=\linewidth]{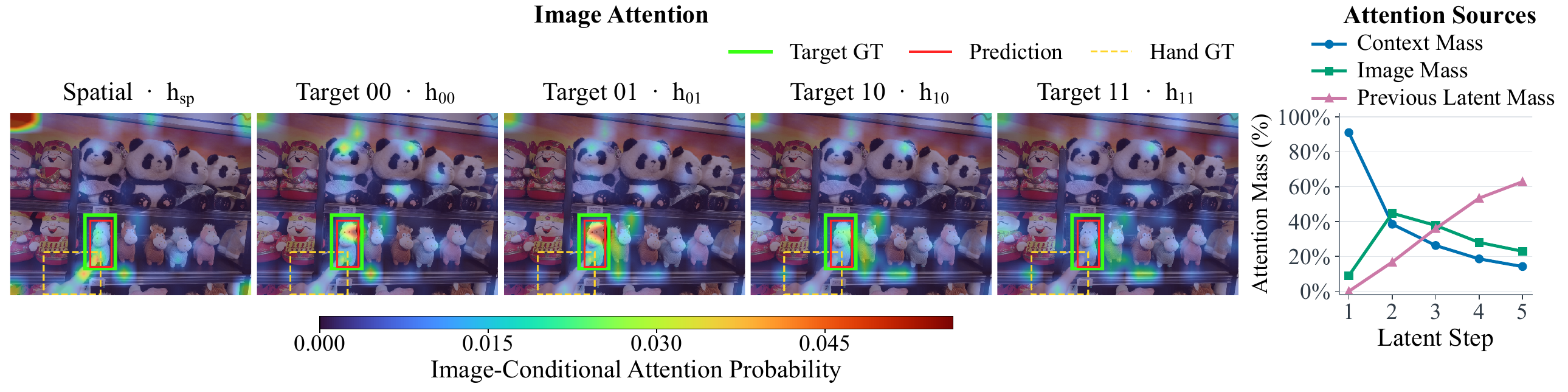}
\caption{Attention across five recurrent latent steps. Left: image-token attention for one pointing example for the spatial state $h_{\mathrm{sp}}$ and target states $h_{00},h_{01},h_{10},h_{11}$, averaged over all 16 heads in layers 20, 24, 28, and 32 and normalized over image tokens using a shared color scale. Green and red boxes mark the ground-truth and predicted target, respectively; the dashed yellow box marks the ground-truth hand region. Right: mean attention mass assigned to context, image tokens, and previous latent states over all evaluation samples.}
\label{fig:latent_attention}
\end{figure}

\paragraph{Attention across latent states.}
Figure~\ref{fig:latent_attention} shows a change in attention allocation across the five latent states. The first spatial state attends predominantly to context. The four subsequent visual states combine attention to context, image tokens, and previous latent states. Image attention peaks at the first visual state, while attention to earlier latent states increases through the sequence. This pattern is consistent with a progression from contextual processing toward integrating visual scene evidence with recurrent context. The heatmaps show one example normalized over image tokens, whereas the curves average source allocation over all evaluation samples; concentrated image attention therefore need not dominate total attention. These views illustrate complementary roles of spatial and visual states, although attention alone does not establish a causal reasoning mechanism.

\section{Conclusion}
\label{sec:conclusion}
SLR organizes pointing geometry and target appearance into recurrent continuous states. Improvements over same-backbone SFT across three backbones and both EgoPoint-Ground hard subsets, alongside positive YouRefIt results, support task-structured intermediate supervision for embodied reference understanding.

\section{Scope and Future Work}
\label{sec:limitations}
Beyond the current single-image, two-dimensional setting, future work includes depth and temporal supervision, weaker annotation requirements, and scene-adaptive states. Controlled state interventions could clarify the reasoning mechanism, while repeated runs and broader evaluations would assess robustness.

\clearpage
\endgroup 
\flushbottom
\subsubsection*{AI Use Statement}
We used large language models to assist with manuscript preparation, including drafting and revising passages, improving grammar and clarity, organizing the presentation of technical material, and refining figure and table captions. The authors are responsible for reviewing and verifying all AI-assisted content and take full responsibility for the accuracy and integrity of the final manuscript.

\subsubsection*{Ethics Statement}
Egocentric imagery can expose people and private environments beyond the intended target. Image use and redistribution must respect dataset access conditions and participant privacy. The photographs and overlays in Figures~\ref{fig:comparison}--\ref{fig:parity} illustrate the method; Figures~\ref{fig:qualitative_comparison} and~\ref{fig:latent_attention} present prediction and attention examples. A predicted region should not be interpreted as a definitive account of a person's intention. Human evaluation requires documented participant and consent procedures.

\subsubsection*{Reproducibility Statement}
Section~\ref{sec:method} specifies the recurrence, supervision targets, and objectives. Appendix~\ref{app:implementation} documents the reference training configuration; Appendices~\ref{app:decoding} and~\ref{app:metrics} specify decoding, metric computation, and timing scope. The available source package contains configuration templates, an evaluation-record merger, source hashes, and contract tests for masking, target construction, and gradient boundaries. It does not include model weights, dataset records, or final per-example results. Reproducing these comparisons also requires split manifests, checkpoints, and run configurations.

\FloatBarrier
\bibliography{iclr2027_conference}
\bibliographystyle{iclr2027_conference}

\clearpage
\appendix
\section*{Appendix}
\input{sections/related_work}

\section{Additional Experimental Details}
\label{app:experimental}
\subsection{Reference Implementation and Training}
\label{app:implementation}
Table~\ref{tab:reference_config} summarizes the exported Exp3.12 configuration for Qwen3.5-4B. It initializes from the base model, freezes the visual encoder and merger, and trains language-side parameters, including token embeddings, together with the spatial head. The existing generic latent markers are retained without role-specific additions. No intermediate textual reasoning steps are used as training targets. The reference configuration disables selector injection and auxiliary bounding-box heads.

\begin{table}[htbp]
\centering
\caption{Reference training configuration for Qwen3.5-4B with a four-epoch training budget. These settings describe the reference implementation rather than all baseline runs.}
\label{tab:reference_config}
\papertablesetup
\begin{tabular}{@{}ll@{}}
\toprule
Setting & Reference value \\
\midrule
Backbone / hidden dimension & Qwen3.5-4B / 2560 \\
Spatial / target states & 1 / 4 \\
Image pixel limits & 50,176 to 262,144 \\
Optimizer / learning rate & AdamW / $10^{-5}$ \\
Weight decay / seed & $0.01$ / 0 \\
AdamW $(\beta_1,\beta_2)$ / $\epsilon$ & $(0.9,0.999)$ / $10^{-8}$ \\
Devices / batch per device / accumulation & 4 / 1 / 3 \\
Training budget & 4 epochs \\
Spatial / target loss weight & $0.005$ / $0.005$ \\
Fingertip / direction coefficient & $0.5$ / $0.5$ \\
SmoothL1 parameter & $\beta=0.1$ \\
Generation limit / cache & 96 tokens / disabled \\
\bottomrule
\end{tabular}
\end{table}

The archived environment is documented as Linux with Python 3.10, PyTorch 2.10.0 (CUDA 12.8), Transformers 5.10.2, and four H100 GPUs. Training uses fully sharded data parallelism with activation checkpointing. Model parameters are converted directly to BF16; the export preserves its BF16 Adam-state strategy without FP32 master weights. AdamW is reinitialized at each epoch, and the Exp3.12 path uses a constant learning rate without a scheduler. Precision, optimizer-state continuity, and scheduling are integral to this recipe; changing any of them alters the training configuration.

The dataset and distributed sampler are shuffled, with the sampler epoch updated each epoch. The nominal effective batch is $4\times1\times3=12$. Checkpoints are saved each epoch as model weights, without a complete optimizer and random-state resume snapshot. Selection uses the validation-generation protocol in Section~\ref{sec:implementation}. The training configuration's separate 48-token field is inactive for online generation evaluation; the exported evaluation worker explicitly uses 96 tokens. Actual post-merge grid dimensions are read from the image processor rather than a fixed input resolution.

The source package includes configuration and source-identity checks and CPU contract tests for the spatial/parity objectives, valid masks, detached targets, and direct gradient paths. These checks establish implementation contracts, not reproduced GPU training or benchmark performance.

\subsection{Bounding-Box Decoding Protocols}
\label{app:decoding}
Free greedy decoding evaluates localization together with structured answer generation. The reference parser accepts an explicit \texttt{bbox\_2d} array or either supported box-tag form, and uses the last structured box in output order. Four unmarked numbers are not accepted as a box. The scorer sorts each pair of horizontal and vertical endpoints, computes continuous-area IoU, and does not clip coordinates to the image range. A missing parse or a nonpositive union yields zero IoU. Parse success identifies a coordinate structure but does not guarantee a valid in-image box.

The tokenizer-aware constrained protocol operates during autoregressive answer generation. Fixed syntax tokens are supplied by the decoder; coordinate values are selected from the model logits using a prefix trie over tokenizations of admissible integers. At each coordinate step, the decoder selects from the allowed next-token set $\mathcal{C}_t(\widehat y_{<t})\subseteq\mathcal{V}$ according to
\begin{equation}
 \widehat y_t=\arg\max_{v\in\mathcal{C}_t(\widehat y_{<t})}
 p_\theta(v\mid I,q,Z_\theta(I,q),\widehat y_{<t}).
\end{equation}
The trie handles multi-token integers using the tokenizer, grammar, and decoded coordinates, without ground-truth boxes. It is a separate control available in the shared implementation; the exported Exp3.12 worker uses free greedy decoding. A matched free/constrained comparison must keep the checkpoint and latent recurrence fixed and report each protocol and its caching settings separately.

\subsection{Metrics, Record Integrity, and Timing}
\label{app:metrics}
For $N$ evaluated examples, let $s_i=\operatorname{IoU}(\widehat b_i,b_i^\star)$, with unparseable predictions assigned zero. We report the following mean overlap and thresholded accuracy
\begin{equation}
 \operatorname{mIoU}=\frac{1}{N}\sum_{i=1}^{N}s_i,
 \qquad
 \operatorname{P}@\tau=\frac{1}{N}\sum_{i=1}^{N}\mathbf{1}[s_i\geq\tau],
 \quad \tau\in\{0.3,0.5,0.7\}.
\end{equation}
All evaluated examples remain in the denominator. The worker requires an annotated target box for every example. Parse failures, end-of-sequence termination, and output-limit hits are logged separately so that changes in answer formatting can be distinguished from localization quality.

The launcher assigns four disjoint shards without padding. Its merger checks complete example coverage, common checkpoint identity, data and record hashes, and unchanged model parameters. Metrics are aggregated over examples rather than obtained by an unweighted average of shard means. The package retains generated token IDs and per-example elapsed time in JSONL records; its summary CSV does not provide mean token count or mean latency.

Generated-token counts can be computed from the stored output IDs, including the terminal token when emitted. They exclude the image/query prefix and its latent carrier and boundary markers. The five recurrent states must therefore be reported separately from output length. The recorded elapsed time starts before sample collation and includes device transfer, provenance hashing, instrumented generation, text decoding, and scoring; it excludes model loading and the initial full-model hash. These instrumented times are not pure decoding latency. A dedicated latency comparison requires a common hardware, batch size, caching, warm-up, synchronization, and timing boundary, with diagnostic overhead removed or reported consistently. Fewer latents need not imply faster inference.

\subsection{Reference Pooling Controls}
\label{app:pooling_controls}
All four reference operators use the same ROI token set and produce four detached targets, with empty groups masked out of the alignment loss. Parity averages each global row/column parity group. Quadrant partitions the ROI tokens at the box midpoint, assigning tokens on the vertical or horizontal dividing line to the left or top half, respectively, and averages each group in top-left, top-right, bottom-left, bottom-right order. Max retains the parity groups and takes an elementwise maximum within each group, isolating the aggregation rule from the spatial partition.

The reference Random operator deterministically permutes ROI token indices by a SHA-256 ordering keyed by the partition version, partition seed, image-relative path, grid dimensions, and ROI-index fingerprint. It splits this ordering into four disjoint groups whose sizes match the corresponding parity groups, and averages each group. For a fixed key and ROI, the partition is repeatable and does not resample during training. This controls the number of tokens per target while changing group membership. These definitions specify the reference controls. The configuration templates do not establish whether the reported Random run used this implementation.

\subsection{Target-Latent Budget Analysis}
\label{app:budget_plot}
\begin{figure}[htbp]
\centering
\includegraphics[width=0.78\linewidth]{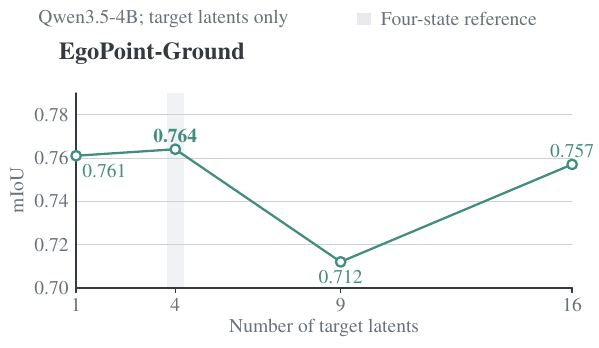}
\caption{Effect of target-state count on EgoPoint-Ground mIoU with Qwen3.5-4B and no spatial state. Points correspond to Table~\ref{tab:ablation_budget}; connecting lines aid visual comparison. Shading marks the four-state reference, and the bold value denotes the highest mIoU. Results are point estimates for the evaluated configurations, with no uncertainty intervals reported.}
\label{fig:latent_budget}
\end{figure}

Figure~\ref{fig:latent_budget} and Table~\ref{tab:ablation_budget} compare four target-state budgets without a spatial state. Four states attain the highest reported mIoU (0.764), closely followed by one state (0.761); one state has higher P@0.7 (0.794 versus 0.783). Increasing the budget to nine or sixteen does not improve mIoU. This sweep supports a compact supervision budget and motivates the four-state reference used in the pooling comparison, without establishing a unique optimum. Changes in target grouping accompany changes in state count, so the comparison concerns the complete supervision configurations.

\begin{table*}[!htb]
\centering
\caption{Ablation of target-state count on \textbf{EgoPoint-Ground} with Qwen3.5-4B and no spatial state. Shading marks the four-state pooling reference; the full model uses five states. Bold values denote the best grounding scores, including ties. See Appendix~\ref{app:metrics} for token-count ($\dagger$) conventions.}
\label{tab:ablation_budget}
\papertablesetup
\begin{tabular*}{\textwidth}{@{\extracolsep{\fill}}llrrrrrr@{}}
\toprule
Backbone & Pooling latents & P@0.3 $\uparrow$ & P@0.5 $\uparrow$ & P@0.7 $\uparrow$ & mIoU $\uparrow$ & Tokens$^\dagger$ & Time (s) $\downarrow$ \\
\midrule
\multicolumn{8}{@{}l}{\textbf{EgoPoint-Ground}} \\
\midrule
\rowcolor{oursrow}
\cellcolor{white}\multirow{4}{*}{Qwen3.5-4B} & \textbf{4 (reference)} & \textbf{0.883} & \textbf{0.856} & 0.783 & \textbf{0.764} & 6.000 & 1.766 \\
 & $1$ & 0.867 & 0.844 & \textbf{0.794} & 0.761 & 3.000 & 1.258 \\
 & $9$ & 0.817 & 0.789 & 0.744 & 0.712 & 11.000 & 2.517 \\
 & $16$ & 0.867 & 0.844 & 0.789 & 0.757 & 18.000 & 3.578 \\
\bottomrule
\end{tabular*}
\end{table*}

\FloatBarrier
\section{Parity-Pooling Properties}
\label{app:parity_properties}
\paragraph{Structural properties.}
The supports satisfy $\Omega_b=\bigsqcup_{\alpha,\beta\in\{0,1\}}\Omega_b^{\alpha\beta}$. In particular, each selected visual token contributes to exactly one descriptor. For a nonempty rectangular support spanning $h$ consecutive rows and $w$ consecutive columns, let $h_\alpha$ and $w_\beta$ count rows and columns of each parity:
\begin{equation}
 n_{\alpha\beta}=h_\alpha w_\beta,\qquad
 h_\alpha\in\{\lfloor h/2\rfloor,\lceil h/2\rceil\},\qquad
 w_\beta\in\{\lfloor w/2\rfloor,\lceil w/2\rceil\}.
 \label{eq:occupancy}
\end{equation}
Hence all four phases are populated when $h,w\geq2$. Thin or unresolved ROIs can leave phases empty; we omit those supervision terms without duplicating tokens, while retaining all four states.

Moreover, the ordinary ROI mean is recovered from the phase means by
\begin{equation}
 \overline V_b=\frac{1}{|\Omega_b|}\sum_{(r,c)\in\Omega_b}V_{r,c}
 =\sum_{\alpha,\beta:n_{\alpha\beta}>0}\frac{n_{\alpha\beta}}{|\Omega_b|}\mu_{\alpha\beta}(V,b).
 \label{eq:mean_recovery}
\end{equation}
Thus, phase means retain the information needed for the single pooled mean, given their counts, while exposing differences between interleaved subsets. Phase-wise averaging is nevertheless lossy: it does not reconstruct the feature grid or guarantee distinct descriptors. Our use of fixed phase order also does not imply translation invariance of the encoder or of the recurrent predictor.

\section{Full Results on the Hard Subsets}
\label{app:hard_results}
Table~\ref{tab:hard_full} expands Table~\ref{tab:main_hard} with all reported methods, the human reference, all localization thresholds, token counts, and times. Hard-Similar (\texttt{valmulti96}) and Hard-Complex (\texttt{valhard96}) are reported separately. The Qwen2.5-VL-7B and Qwen3-VL-8B Standard SFT rows use run \texttt{\_02}. $\dagger$ denotes the reported token counts. Appendix~\ref{app:metrics} discusses timing scope and token-count conventions for interpreting the reported token counts and elapsed times.

\begingroup
\papertablesetup
\setlength{\LTleft}{0pt}
\setlength{\LTright}{0pt}
\setlength{\LTcapwidth}{\textwidth}
\setlength{\LTpre}{6pt}
\setlength{\LTpost}{6pt}
\begin{longtable}{@{\extracolsep{\fill}}llrrrrrr@{}}
\caption{\normalsize Full results on the \textbf{EgoPoint-Ground} Hard-Similar and Hard-Complex subsets. Shading identifies SLR; bold values denote the best grounding scores within each backbone and subset, including ties. Dashes indicate unreported values; N/A indicates inapplicable entries.}\label{tab:hard_full} \\
\toprule
Model & Method & P@0.3 $\uparrow$ & P@0.5 $\uparrow$ & P@0.7 $\uparrow$ & mIoU $\uparrow$ & Tokens$^\dagger$ & Time (s) $\downarrow$ \\
\midrule
\endfirsthead
\multicolumn{8}{@{}l}{\small Table~\thetable{} (continued)} \\
\toprule
Model & Method & P@0.3 $\uparrow$ & P@0.5 $\uparrow$ & P@0.7 $\uparrow$ & mIoU $\uparrow$ & Tokens$^\dagger$ & Time (s) $\downarrow$ \\
\midrule
\endhead
\midrule
\multicolumn{8}{r@{}}{\footnotesize Continued on next page.} \\
\endfoot
\bottomrule
\endlastfoot
\multicolumn{8}{@{}l}{\textbf{Hard-Similar: multiple visually similar objects}} \\*
Human Study & Human Participants & 0.965 & 0.948 & 0.917 & 0.856 & N/A & N/A \\*
\addlinespace[2pt]
\multirow{5}{*}{Qwen2.5-VL-7B} & Zero-shot & 0.281 & 0.240 & 0.167 & 0.214 & 0.000 & 0.864 \\*
 & Standard SFT & 0.281 & 0.229 & 0.146 & 0.202 & 0.000 & 0.576 \\*
 & Text CoT (adapted) & 0.292 & 0.188 & 0.063 & 0.198 & 166.854 & 3.312 \\*
 & PointVG-R & \textbf{0.354} & 0.313 & 0.177 & 0.276 & 175.521 & 3.511 \\*
\rowcolor{oursrow}
\cellcolor{white} & \textbf{SLR} & \textbf{0.354} & \textbf{0.323} & \textbf{0.313} & \textbf{0.327} & 7.000 & 0.755 \\
\addlinespace[2pt]
\multirow{4}{*}{Qwen3-VL-8B} & Zero-shot & 0.271 & 0.271 & 0.271 & 0.271 & 0.000 & 1.111 \\*
 & Standard SFT & 0.292 & 0.281 & 0.281 & 0.290 & 0.000 & 1.074 \\*
 & Text CoT (adapted) & \textbf{0.375} & \textbf{0.333} & \textbf{0.323} & \textbf{0.353} & 179.302 & 4.473 \\*
\rowcolor{oursrow}
\cellcolor{white} & \textbf{SLR} & 0.313 & 0.302 & 0.292 & 0.315 & 7.000 & 0.884 \\
\addlinespace[2pt]
\multirow{5}{*}{Qwen3.5-4B} & Zero-shot (Non-thinking) & 0.198 & 0.167 & 0.156 & 0.161 & 0.000 & 1.238 \\*
& Zero-shot (Thinking) & 0.271 & 0.240 & 0.188 & 0.228 & 130.615 & 5.149 \\*
 & Standard SFT & 0.417 & 0.396 & 0.344 & 0.361 & 0.000 & 1.101 \\*
 & LVR & 0.177 & 0.167 & 0.135 & 0.158 & 6.000 & 1.329 \\*
\rowcolor{oursrow}
\cellcolor{white} & \textbf{SLR} & \textbf{0.500} & \textbf{0.458} & \textbf{0.396} & \textbf{0.421} & 7.000 & 1.899 \\
\midrule
\multicolumn{8}{@{}l}{\textbf{Hard-Complex: complex scenes}} \\*
Human Study & Human Participants & 0.958 & 0.920 & 0.809 & 0.796 & N/A & N/A \\*
\addlinespace[2pt]
\multirow{5}{*}{Qwen2.5-VL-7B} & Zero-shot & 0.531 & 0.490 & 0.396 & 0.416 & 0.000 & 0.869 \\*
 & Standard SFT & 0.510 & 0.458 & 0.323 & 0.389 & 0.000 & 0.564 \\*
 & Text CoT (adapted) & 0.583 & 0.448 & 0.208 & 0.411 & 168.760 & 3.368 \\*
 & PointVG-R & \textbf{0.656} & 0.573 & 0.365 & 0.473 & 178.938 & 3.535 \\*
\rowcolor{oursrow}
\cellcolor{white} & \textbf{SLR} & 0.615 & \textbf{0.583} & \textbf{0.469} & \textbf{0.511} & 7.000 & 0.748 \\
\addlinespace[2pt]
\multirow{4}{*}{Qwen3-VL-8B} & Zero-shot & 0.427 & 0.396 & 0.302 & 0.354 & 0.000 & 1.116 \\*
 & Standard SFT & 0.427 & 0.385 & 0.313 & 0.349 & 0.000 & 1.095 \\*
 & Text CoT (adapted) & 0.531 & 0.500 & 0.406 & 0.452 & 180.333 & 4.676 \\*
\rowcolor{oursrow}
\cellcolor{white} & \textbf{SLR} & \textbf{0.583} & \textbf{0.542} & \textbf{0.448} & \textbf{0.495} & 7.000 & 0.930 \\
\addlinespace[2pt]
\multirow{5}{*}{Qwen3.5-4B} & Zero-shot (Non-thinking) & 0.323 & 0.302 & 0.260 & 0.274 & 0.000 & 1.352 \\*
& Zero-shot (Thinking) & 0.375 & 0.333 & 0.240 & 0.293 & 150.188 & 5.629 \\*
 & Standard SFT & 0.667 & 0.594 & 0.458 & 0.536 & 0.000 & 1.145 \\*
 & LVR & 0.500 & 0.396 & 0.333 & 0.396 & 6.000 & 1.308 \\*
\rowcolor{oursrow}
\cellcolor{white} & \textbf{SLR} & \textbf{0.719} & \textbf{0.656} & \textbf{0.479} & \textbf{0.551} & 7.000 & 1.889 \\
\end{longtable}
\endgroup

\clearpage
\section{Supplementary Computation Records}
\label{app:computation_records}
Table~\ref{tab:computation_records} retains the token counts and times accompanying the main comparisons and ablations. The values are reproduced without converting their units. In particular, the reported zero counts for direct prediction and the fixed counts for latent methods are not total answer lengths. Continuous states, boundary markers, and generated text require separate accounting; these aggregate records do not establish a common counting convention across method families. Timing scope is discussed in Appendix~\ref{app:metrics}. Localization results in the main text do not rely on an efficiency ranking.

\begin{table}[!htb]
\centering
\caption{Reported token counts and elapsed times for Tables~\ref{tab:main_egopoint}, \ref{tab:ablation_components}, and~\ref{tab:ablation_pooling}. Appendix~\ref{app:metrics} details token-count ($\dagger$) and timing conventions. Standard denotes EgoPoint-Ground.}
\label{tab:computation_records}
\papertablesetup
\begin{tabular*}{\textwidth}{@{\extracolsep{\fill}}lllrr@{}}
\toprule
Backbone & Method / variant & Evaluation set & Tokens$^\dagger$ & Time (s) \\
\midrule
\multicolumn{5}{@{}l}{\textbf{Main comparisons}} \\
Qwen2.5-VL-7B & Zero-shot & Standard & 0.000 & 1.181 \\
 & Standard SFT & Standard & 0.000 & 0.885 \\
 & Text CoT (adapted) & Standard & 173.533 & 4.402 \\
 & PointVG-R & Standard & 183.622 & 4.708 \\
 & SLR & Standard & 7.000 & 1.362 \\
Qwen3-VL-8B & Zero-shot & Standard & 0.000 & 1.444 \\
 & Standard SFT & Standard & 0.000 & 1.421 \\
 & Text CoT (adapted) & Standard & 176.006 & 5.602 \\
 & SLR & Standard & 7.000 & 1.448 \\
Qwen3.5-4B & Zero-shot (Non-thinking) & Standard & 0.000 & 1.424 \\
 & Zero-shot (Thinking) & Standard & 150.267 & 5.685 \\
 & Standard SFT & Standard & 0.000 & 1.118 \\
 & LVR & Standard & 6.000 & 1.341 \\
 & SLR & Standard & 7.000 & 1.859 \\
\midrule
\multicolumn{5}{@{}l}{\textbf{Component configurations}} \\
Qwen3.5-4B & Spatial + Parity (SLR) & Standard & 7.000 & 1.859 \\
 & Spatial-only & Standard & 3.000 & 1.268 \\
 & Target-only & Standard & 6.000 & 1.766 \\
 & SFT & Standard & 0.000 & 1.118 \\
 & Spatial + Parity (SLR) & Hard-Similar & 7.000 & 1.899 \\
 & Spatial-only & Hard-Similar & 3.000 & 1.276 \\
 & Target-only & Hard-Similar & 6.000 & 1.753 \\
 & SFT & Hard-Similar & 0.000 & 1.101 \\
 & Spatial + Parity (SLR) & Hard-Complex & 7.000 & 1.889 \\
 & Spatial-only & Hard-Complex & 3.000 & 1.243 \\
 & Target-only & Hard-Complex & 6.000 & 1.743 \\
 & SFT & Hard-Complex & 0.000 & 1.145 \\
\midrule
\multicolumn{5}{@{}l}{\textbf{Pooling operators}} \\
Qwen3.5-4B & Parity & Standard & 6.000 & 1.766 \\
 & Quadrant & Standard & 6.000 & 1.754 \\
 & Random & Standard & 6.000 & 1.741 \\
 & Max & Standard & 6.000 & 1.740 \\
\bottomrule
\end{tabular*}
\end{table}

\end{document}

%% file: math_commands.tex
\usepackage{amsmath,amsfonts,bm}

\def\eqref#1{equation~\ref{#1}}

\def\1{\bm{1}}

\DeclareMathAlphabet{\mathsfit}{\encodingdefault}{\sfdefault}{m}{sl}
\SetMathAlphabet{\mathsfit}{bold}{\encodingdefault}{\sfdefault}{bx}{n}



%% file: sections/related_work.tex
\section{Related Work}
\label{app:related}
\label{sec:related}

\subsection{Referring Expressions and Embodied Pointing}
Referring expression comprehension localizes the object specified by a language expression. MAttNet~\citep{yu2018mattnet} separates subject appearance, location, and inter-object relationships into modules whose contributions depend on the expression. TransVG~\citep{deng2021transvg} instead uses Transformer-based multimodal fusion and directly regresses bounding-box coordinates. These approaches establish that localization depends on both object content and its relation to the surrounding scene. Pointing introduces a further source of reference: the gesture can disambiguate objects whose appearance and linguistic descriptions are similar.

YouRefIt~\citep{chen2021yourefit} explicitly studies embodied reference understanding with language and gesture, including the perspective differences between the person making a reference and the observer. EgoPoint~\citep{li2026egopoint} develops egocentric hand-pointing data and a structured visual chain-of-thought baseline. PointVG-R~\citep{li2026pointvgr} pursues precise pointing localization through geometric reasoning traces, supervised fine-tuning, and reinforcement learning. Together, these works motivate treating pointing as a relation between a gesture and a candidate referent, rather than identifying the hand or recognizing the target category alone. Our setting follows this grounding task, but studies how to supervise the model's recurrent continuous states. The fingertip and root-to-tip direction supervise a two-dimensional ray representation. The box is predicted from the image, query, and latent sequence, without a prescribed ray--object intersection rule.

\subsection{Grounded Multimodal Models and Spatial Knowledge}
Grounded multimodal language models connect linguistic representations to image regions. Kosmos-2~\citep{peng2024kosmos2} associates text spans with bounding boxes and represents location information in its output sequence. Ferret~\citep{you2024ferret} combines discrete spatial coordinates with continuous visual features to represent regions at different granularities. These models demonstrate the value of spatially indexed features. Prompted regions and output coordinates serve different roles from recurrent hidden states generated before the answer.

SpatialVLM~\citep{chen2024spatialvlm} improves qualitative and quantitative spatial reasoning through automatically generated spatial question--answer data. SpatialRGPT~\citep{cheng2024spatialrgpt} develops grounded spatial reasoning using region information, a spatial data pipeline, and optional depth features. Their emphasis is learning spatial knowledge and answering spatial questions, including relations in three dimensions. Our geometry supervision is narrower: a fingertip position and an image-grid-scaled direction specify a two-dimensional pointing state. This auxiliary objective does not reconstruct the three-dimensional scene or uniquely determine the referred object. It is combined with target-region visual supervision because gesture geometry and candidate appearance supply different information for the grounding decision.

\subsection{Explicit Visual Reasoning and Thinking with Images}
Chain-of-thought prompting makes intermediate reasoning explicit in text~\citep{wei2022cot}. Multimodal extensions can also expose visual evidence or manipulate images during reasoning. V*~\citep{wu2024vstar} uses guided visual search and visual working memory to acquire fine-grained evidence. Visual Sketchpad~\citep{hu2024sketchpad} enables models to draw and use visual tools, while DeepEyes~\citep{zheng2025deepeyes} trains image-tool use with reinforcement learning. We use \emph{thinking with images} as a family-level description for such explicit visual interactions; Figure~\ref{fig:comparison} depicts a representative crop-and-inspect path, not a shared implementation of all these methods.

A complementary line of work grounds intermediate rationales in image regions. Visual CoT~\citep{shao2024visualcot} provides bounding-box annotations for intermediate visual reasoning and supports iterative attention to relevant regions. Grounded Chain-of-Thought~\citep{wu2026gcot} connects reasoning steps to localized visual cues and evaluates answer--grounding consistency. \citet{xia2025groundedcot} bootstrap grounded reasoning traces for data-efficient model adaptation. PointVG-R~\citep{li2026pointvgr} specializes explicit geometric reasoning to pointing localization. These approaches expose intermediate evidence and incorporate geometry into explicit reasoning.

The relevant design question is which intermediate representation is useful for the task. In controlled vision-centric generalization experiments, \citet{du2026lengthy} find that concise grounded reasoning can be more effective than lengthy textual chains. This supports examining reasoning format and budget, but does not establish that text is universally inadequate or that continuous states are necessary. Our model uses recurrent hidden-state feedback without intermediate textual rationales or visual-tool calls. Any accuracy or latency advantage over explicit reasoning must nevertheless be measured under matched evaluation conditions; shorter output omits recurrent computation costs.

\subsection{Continuous and Latent Visual Reasoning}
Coconut~\citep{hao2024coconut} provides the central continuous transition used here: a hidden state is fed back as the next input embedding without first selecting a vocabulary token. SoftCoT~\citep{xu2025softcot} offers a different route to soft reasoning representations, using an assistant model and a learned projection to supply soft thoughts to a language model. These methods motivate embedding-space computation without specifying the gesture geometry or target appearance to encode.

LVR~\citep{li2025lvr} directly addresses visual grounding of latent computation by aligning predicted continuous states with visual embeddings from annotated regions. It is the closest algorithmic foundation for our target supervision, and already supports reasoning with a bounded number of visual latents. ILVR~\citep{dong2026ilvr} interleaves textual and latent reasoning and selectively models perceptual features from intermediate images. LIVR~\citep{li2026livr} uses learned latent tokens and an attention-bottleneck training stage to encourage implicit visual processing without intermediate visual annotations. Its learned prompt tokens differ from the freely generated hidden vectors recursively fed back in Coconut and our model. ATLAS~\citep{guo2026atlas} also internalizes visual operations, but uses functional vocabulary tokens rather than the same full-dimensional continuous recurrence. Thus, ``latent visual reasoning'' encompasses distinct mechanisms and supervision.

Especially relevant is RIS~\citep{cui2026ris}, which grounds latent reasoning in both spatial and semantic information. RIS uses auxiliary box and semantic prediction heads, a curriculum from explicit grounded steps to latent reasoning, and a later attention-masking stage. Its reported setup also uses five latent tokens. RIS therefore provides a direct precedent for spatial-semantic latent supervision and auxiliary readouts under a compact reasoning budget. Our design assigns task-specific supervision to an ordered recurrent sequence: one state predicts fingertip position and pointing direction, followed by four states aligned with parity-pooled target features. It does not require intermediate textual reasoning traces or region-description targets. Our focus is this task-specific factorization of geometry and visual evidence within freely generated recurrent states.

Across these comparisons, our auxiliary labels constrain what information is recoverable from the recurrent states. They are not substituted for those states, and are not available at inference. Successful decoding of geometry is evidence of representation content; demonstrating that the box prediction depends on that geometry requires additional interventions or controlled ablations.

\subsection{Region Pooling and Polyphase Representations}
Region pooling converts a variable spatial support into a fixed-size representation. Spatial pyramid pooling~\citep{he2014spp} aggregates features in spatial bins at multiple resolutions. RoIAlign, introduced in Mask R-CNN~\citep{he2017maskrcnn}, avoids quantizing region boundaries and uses bilinear interpolation to improve feature alignment. These methods motivate a general distinction between the region being represented and the operator used to summarize its features. Parity pooling uses post-merge image tokens within the annotated box, without cropping or re-encoding.

The four groups arise from a classical signal-processing construction. Polyphase decomposition partitions a discrete signal into interleaved subsequences indexed by sampling phase~\citep{smith2011spectral}. In two dimensions, the row and column residues modulo two define four interleaved lattices. Work on convolutional downsampling further studies how sampling phase affects shift sensitivity: \citet{zhang2019shift} applies low-pass filtering before subsampling, and \citet{chaman2021shift} develops adaptive polyphase sampling for shift-invariant networks. These are precedents for reasoning about sampling phase, without implying that all uses of parity groups inherit these invariance properties.

We transfer the four-phase lattice partition to the construction of visual supervision targets. A mean is computed over each parity class within the target support, with phase indices anchored to the full image grid. Unlike four quadrants, the classes are spatially interleaved instead of representing contiguous parts of the box. Unlike one region mean, they retain four separate phase-conditioned summaries. The polyphase partition itself is established, and averaging each class is lossy: our construction does not imply invertibility, anti-aliasing, or translation invariance. Our contribution concerns the use of these phase-conditioned visual targets alongside a supervised ray state in recurrent pointing-grounding computation. Comparisons with quadrant means, count-matched random-group means, and maxima on parity supports test the grouping and aggregation choices. Their relative performance is an empirical question, rather than a consequence of the signal-processing analogy.

\subsection{Autoregressive Formulation and LVR Supervision}
\label{app:latent_background}
For a multimodal prompt and model-generated states $Z_\theta(I,q)$, answer generation factorizes as
\begin{equation}
 p_\theta(y\mid I,q)=\prod_{t=1}^{|y|}
 p_\theta\!\left(y_t\mid I,q,Z_\theta(I,q),y_{<t}\right).
 \label{eq:conditional}
\end{equation}

\paragraph{Visual supervision in LVR.}
LVR~\citep{li2025lvr} gives continuous states a visual training target. An annotated ROI selects projected image tokens $(v_1,\ldots,v_M)$. Its supervised stage places these tokens in the latent segment as teacher-forced inputs and trains the preceding hidden states $\widetilde h_j$ to reconstruct them, jointly with next-token prediction:
\begin{equation}
 \mathcal{L}_{\mathrm{LVR}}^{\mathrm{SFT}}
 =\mathcal{L}_{\mathrm{NTP}}+
 \frac{\lambda_{\mathrm{LVR}}}{M}\sum_{j=1}^{M}
 \|\widetilde h_j-\operatorname{sg}(v_j)\|_2^2.
 \label{eq:lvr_foundation_loss}
\end{equation}
At inference, LVR feeds predicted hidden vectors back between latent boundary markers.

%% file: iclr2027_conference.bib
@inproceedings{yu2018mattnet,
  title={{MAttNet}: Modular Attention Network for Referring Expression Comprehension},
  author={Yu, Licheng and Lin, Zhe and Shen, Xiaohui and Yang, Jimei and Lu, Xin and Bansal, Mohit and Berg, Tamara L.},
  booktitle={Proceedings of the IEEE Conference on Computer Vision and Pattern Recognition},
  pages={1307--1315},
  year={2018},
  url={https://openaccess.thecvf.com/content_cvpr_2018/html/Yu_MAttNet_Modular_Attention_CVPR_2018_paper.html}
}

@inproceedings{deng2021transvg,
  title={{TransVG}: End-to-End Visual Grounding With Transformers},
  author={Deng, Jiajun and Yang, Zhengyuan and Chen, Tianlang and Zhou, Wengang and Li, Houqiang},
  booktitle={Proceedings of the IEEE/CVF International Conference on Computer Vision},
  pages={1769--1779},
  year={2021},
  url={https://openaccess.thecvf.com/content/ICCV2021/html/Deng_TransVG_End-to-End_Visual_Grounding_With_Transformers_ICCV_2021_paper.html}
}

@inproceedings{chen2021yourefit,
  title={{YouRefIt}: Embodied Reference Understanding With Language and Gesture},
  author={Chen, Yixin and Li, Qing and Kong, Deqian and Kei, Yik Lun and Zhu, Song-Chun and Gao, Tao and Zhu, Yixin and Huang, Siyuan},
  booktitle={Proceedings of the IEEE/CVF International Conference on Computer Vision},
  pages={1385--1395},
  year={2021},
  url={https://openaccess.thecvf.com/content/ICCV2021/html/Chen_YouRefIt_Embodied_Reference_Understanding_With_Language_and_Gesture_ICCV_2021_paper.html}
}

@inproceedings{shi2022rep,
  title={Spatial and Visual Perspective-Taking via View Rotation and Relation Reasoning for Embodied Reference Understanding},
  author={Shi, Cheng and Yang, Sibei},
  booktitle={European Conference on Computer Vision},
  year={2022},
  url={https://www.ecva.net/papers/eccv_2022/papers_ECCV/papers/136960199.pdf}
}

@inproceedings{li2023touchline,
  title={Understanding Embodied Reference with Touch-Line Transformer},
  author={Li, Yang and Chen, Xiaoxue and Zhao, Hao and Gong, Jiangtao and Zhou, Guyue and Rossano, Federico and Zhu, Yixin},
  booktitle={International Conference on Learning Representations},
  year={2023},
  url={https://yzhu.io/publication/pointing2023iclr/}
}

@article{guo2024addino,
  title={{AD-DINO}: Attention-Dynamic {DINO} for Distance-Aware Embodied Reference Understanding},
  author={Guo, Hao and Fan, Wei and Wei, Baichun and Zhu, Jianfei and Tian, Jin and Yi, Chunzhi and Jiang, Feng},
  journal={arXiv preprint arXiv:2411.08451},
  year={2024},
  url={https://arxiv.org/abs/2411.08451}
}

@inproceedings{eyiokur2026cape,
  title={{CAPE}: A {CLIP}-Aware Pointing Ensemble of Complementary Heatmap Cues for Embodied Reference Understanding},
  author={Eyiokur, Fevziye Irem and Yaman, Dogucan and Ekenel, Haz{\i}m Kemal and Waibel, Alexander},
  booktitle={Proceedings of the IEEE/CVF Winter Conference on Applications of Computer Vision},
  pages={3939--3950},
  year={2026},
  url={https://openaccess.thecvf.com/content/WACV2026/html/Eyiokur_CAPE_A_CLIP-Aware_Pointing_Ensemble_of_Complementary_Heatmap_Cues_for_WACV_2026_paper.html}
}

@inproceedings{eyiokur2026daeru,
  title={A Multimodal Depth-Aware Method for Embodied Reference Understanding},
  author={Eyiokur, Fevziye Irem and Yaman, Dogucan and Ekenel, Haz{\i}m Kemal and Waibel, Alexander},
  booktitle={IEEE International Conference on Acoustics, Speech and Signal Processing},
  year={2026},
  url={https://arxiv.org/abs/2510.08278}
}

@inproceedings{li2026egopoint,
  title={Beyond Language: Grounding Referring Expressions with Hand Pointing in Egocentric Vision},
  author={Li, Ling and Liu, Bowen and Zhan, Zinuo and Jie, Peng and Zhong, Jianhui and Chang, Kenglun and Deng, Zhidong},
  booktitle={European Conference on Computer Vision},
  year={2026}
}

@article{li2026pointvgr,
  title={{PointVG-R}: Internalizing Geometric Reasoning in {MLLMs} for Precise Pointing Localization via Visual Chain of Thought},
  author={Li, Ling and Liu, Bowen and Zhan, Zinuo and Zhong, Jianhui and Zhu, Ziyu and Wei, Bingcai and Chang, Kenglun and Deng, Zhidong},
  journal={arXiv preprint arXiv:2606.24539},
  year={2026},
  url={https://arxiv.org/abs/2606.24539}
}

@inproceedings{peng2024kosmos2,
  title={{Kosmos-2}: Grounding Multimodal Large Language Models to the World},
  author={Peng, Zhiliang and Wang, Wenhui and Dong, Li and Hao, Yaru and Huang, Shaohan and Ma, Shuming and Wei, Furu},
  booktitle={International Conference on Learning Representations},
  year={2024},
  url={https://openreview.net/forum?id=lLmqxkfSIw}
}

@inproceedings{you2024ferret,
  title={{Ferret}: Refer and Ground Anything Anywhere at Any Granularity},
  author={You, Haoxuan and Zhang, Haotian and Gan, Zhe and Du, Xianzhi and Zhang, Bowen and Wang, Zirui and Cao, Liangliang and Chang, Shih-Fu and Yang, Yinfei},
  booktitle={International Conference on Learning Representations},
  year={2024},
  url={https://arxiv.org/abs/2310.07704}
}

@inproceedings{chen2024spatialvlm,
  title={{SpatialVLM}: Endowing Vision-Language Models with Spatial Reasoning Capabilities},
  author={Chen, Boyuan and Xu, Zhuo and Kirmani, Sean and Ichter, Brian and Sadigh, Dorsa and Guibas, Leonidas and Xia, Fei},
  booktitle={Proceedings of the IEEE/CVF Conference on Computer Vision and Pattern Recognition},
  pages={14455--14465},
  year={2024},
  url={https://openaccess.thecvf.com/content/CVPR2024/html/Chen_SpatialVLM_Endowing_Vision-Language_Models_with_Spatial_Reasoning_Capabilities_CVPR_2024_paper.html}
}

@inproceedings{cheng2024spatialrgpt,
  title={{SpatialRGPT}: Grounded Spatial Reasoning in Vision-Language Models},
  author={Cheng, An-Chieh and Yin, Hongxu and Fu, Yang and Guo, Qiushan and Yang, Ruihan and Kautz, Jan and Wang, Xiaolong and Liu, Sifei},
  booktitle={Advances in Neural Information Processing Systems},
  volume={37},
  year={2024},
  url={https://proceedings.neurips.cc/paper_files/paper/2024/hash/f38cb4cf9a5eaa92b3cfa481832719c6-Abstract-Conference.html}
}

@inproceedings{wei2022cot,
  title={Chain-of-Thought Prompting Elicits Reasoning in Large Language Models},
  author={Wei, Jason and Wang, Xuezhi and Schuurmans, Dale and Bosma, Maarten and Ichter, Brian and Xia, Fei and Chi, Ed H. and Le, Quoc V. and Zhou, Denny},
  booktitle={Advances in Neural Information Processing Systems},
  volume={35},
  year={2022},
  url={https://proceedings.neurips.cc/paper/2022/hash/9d5609613524ecf4f15af0f7b31abca4-Abstract-Conference.html}
}

@inproceedings{wu2024vstar,
  title={{V*}: Guided Visual Search as a Core Mechanism in Multimodal {LLMs}},
  author={Wu, Penghao and Xie, Saining},
  booktitle={Proceedings of the IEEE/CVF Conference on Computer Vision and Pattern Recognition},
  pages={13084--13094},
  year={2024},
  url={https://openaccess.thecvf.com/content/CVPR2024/html/Wu_V_Guided_Visual_Search_as_a_Core_Mechanism_in_Multimodal_CVPR_2024_paper.html}
}

@inproceedings{hu2024sketchpad,
  title={Visual Sketchpad: Sketching as a Visual Chain of Thought for Multimodal Language Models},
  author={Hu, Yushi and Shi, Weijia and Fu, Xingyu and Roth, Dan and Ostendorf, Mari and Zettlemoyer, Luke and Smith, Noah A. and Krishna, Ranjay},
  booktitle={Advances in Neural Information Processing Systems},
  volume={37},
  year={2024},
  url={https://proceedings.neurips.cc/paper_files/paper/2024/hash/fb82011040977c7712409fbdb5456647-Abstract-Conference.html}
}

@article{zheng2025deepeyes,
  title={{DeepEyes}: Incentivizing ``Thinking with Images'' via Reinforcement Learning},
  author={Zheng, Ziwei and Yang, Michael and Hong, Jack and Zhao, Chenxiao and Xu, Guohai and Yang, Le and Shen, Chao and Yu, Xing},
  journal={arXiv preprint arXiv:2505.14362},
  year={2025},
  url={https://arxiv.org/abs/2505.14362}
}

@inproceedings{shao2024visualcot,
  title={Visual {CoT}: Advancing Multi-Modal Language Models with a Comprehensive Dataset and Benchmark for Chain-of-Thought Reasoning},
  author={Shao, Hao and Qian, Shengju and Xiao, Han and Song, Guanglu and Zong, Zhuofan and Wang, Letian and Liu, Yu and Li, Hongsheng},
  booktitle={Advances in Neural Information Processing Systems},
  volume={37},
  year={2024},
  url={https://proceedings.neurips.cc/paper_files/paper/2024/hash/0ff38d72a2e0aa6dbe42de83a17b2223-Abstract-Datasets_and_Benchmarks_Track.html}
}

@inproceedings{wu2026gcot,
  title={Grounded Chain-of-Thought for Multimodal Large Language Models},
  author={Wu, Qiong and Yang, Xiangcong and Zhou, Yiyi and Fang, Chenxin and Song, Baiyang and Sun, Xiaoshuai and Ji, Rongrong},
  booktitle={Proceedings of the IEEE/CVF Conference on Computer Vision and Pattern Recognition},
  pages={33577--33587},
  year={2026},
  url={https://openaccess.thecvf.com/content/CVPR2026/html/Wu_Grounded_Chain-of-Thought_for_Multimodal_Large_Language_Models_CVPR_2026_paper.html}
}

@inproceedings{xia2025groundedcot,
  title={Bootstrapping Grounded Chain-of-Thought in Multimodal {LLMs} for Data-Efficient Model Adaptation},
  author={Xia, Jiaer and Tong, Bingkui and Zang, Yuhang and Shao, Rui and Zhou, Kaiyang},
  booktitle={Proceedings of the IEEE/CVF International Conference on Computer Vision},
  pages={208--217},
  year={2025},
  url={https://openaccess.thecvf.com/content/ICCV2025/html/Xia_Bootstrapping_Grounded_Chain-of-Thought_in_Multimodal_LLMs_for_Data-Efficient_Model_Adaptation_ICCV_2025_paper.html}
}

@inproceedings{du2026lengthy,
  title={Revisiting the Necessity of Lengthy Chain-of-Thought in Vision-centric Reasoning Generalization},
  author={Du, Yifan and Zhou, Kun and Min, Yingqian and Ling, Yue and Zhao, Wayne Xin and Wu, Youbin and Wen, Ji-Rong},
  booktitle={Proceedings of the IEEE/CVF Conference on Computer Vision and Pattern Recognition},
  pages={12019--12029},
  year={2026},
  url={https://openaccess.thecvf.com/content/CVPR2026/html/Du_Revisiting_the_Necessity_of_Lengthy_Chain-of-Thought_in_Vision-centric_Reasoning_Generalization_CVPR_2026_paper.html}
}

@inproceedings{hao2024coconut,
  title={Training Large Language Models to Reason in a Continuous Latent Space},
  author={Hao, Shibo and Sukhbaatar, Sainbayar and Su, DiJia and Li, Xian and Hu, Zhiting and Weston, Jason and Tian, Yuandong},
  booktitle={Conference on Language Modeling},
  year={2025},
  url={https://arxiv.org/abs/2412.06769}
}

@inproceedings{xu2025softcot,
  title={{SoftCoT}: Soft Chain-of-Thought for Efficient Reasoning with {LLMs}},
  author={Xu, Yige and Guo, Xu and Zeng, Zhiwei and Miao, Chunyan},
  booktitle={Proceedings of the 63rd Annual Meeting of the Association for Computational Linguistics (Volume 1: Long Papers)},
  pages={23336--23351},
  year={2025},
  url={https://aclanthology.org/2025.acl-long.1137/}
}

@inproceedings{li2025lvr,
  title={Latent Visual Reasoning},
  author={Li, Bangzheng and Sun, Ximeng and Liu, Jiang and Wang, Ze and Wu, Jialian and Yu, Xiaodong and Chen, Hao and Barsoum, Emad and Chen, Muhao and Liu, Zicheng},
  booktitle={International Conference on Learning Representations},
  year={2026},
  url={https://proceedings.iclr.cc/paper_files/paper/2026/file/efa0939caa8a906fd0a61a9c60e46e8b-Paper-Conference.pdf}
}

@inproceedings{dong2026ilvr,
  title={Interleaved Latent Visual Reasoning with Selective Perceptual Modeling},
  author={Dong, Shuai and Wang, Siyuan and Liu, Xingyu and Li, Chenglin and Hou, Haowen and Wei, Zhongyu},
  booktitle={Proceedings of the 64th Annual Meeting of the Association for Computational Linguistics (Volume 1: Long Papers)},
  pages={29316--29335},
  year={2026},
  url={https://aclanthology.org/2026.acl-long.1351/}
}

@inproceedings{li2026livr,
  title={Latent Implicit Visual Reasoning},
  author={Li, Kelvin and Shang, Chuyi and Karlinsky, Leonid and Feris, Rogerio and Darrell, Trevor and Herzig, Roei},
  booktitle={Proceedings of the IEEE/CVF Conference on Computer Vision and Pattern Recognition},
  pages={33457--33466},
  year={2026},
  url={https://openaccess.thecvf.com/content/CVPR2026/html/Li_Latent_Implicit_Visual_Reasoning_CVPR_2026_paper.html}
}

@article{guo2026atlas,
  title={{ATLAS}: Agentic or Latent Visual Reasoning? One Word is Enough for Both},
  author={Guo, Ziyu and Liu, Rain and Chen, Xinyan and Heng, Pheng-Ann},
  journal={arXiv preprint arXiv:2605.15198},
  year={2026},
  url={https://arxiv.org/abs/2605.15198}
}

@article{cui2026ris,
  title={Retrieve, Integrate, and Synthesize: Spatial-Semantic Grounded Latent Visual Reasoning},
  author={Cui, Jin and Long, Xinyue and Zhang, Xunyong and Zhang, Yadong and Su, Chuanchang and Gan, Jingye and Zhao, Boran and Ren, Pengju},
  journal={arXiv preprint arXiv:2605.07106},
  year={2026},
  url={https://arxiv.org/abs/2605.07106}
}

@inproceedings{he2014spp,
  title={Spatial Pyramid Pooling in Deep Convolutional Networks for Visual Recognition},
  author={He, Kaiming and Zhang, Xiangyu and Ren, Shaoqing and Sun, Jian},
  booktitle={European Conference on Computer Vision},
  pages={346--361},
  year={2014},
  url={https://arxiv.org/abs/1406.4729}
}

@inproceedings{he2017maskrcnn,
  title={{Mask R-CNN}},
  author={He, Kaiming and Gkioxari, Georgia and Doll{\'a}r, Piotr and Girshick, Ross},
  booktitle={Proceedings of the IEEE International Conference on Computer Vision},
  pages={2961--2969},
  year={2017},
  url={https://openaccess.thecvf.com/content_iccv_2017/html/He_Mask_R-CNN_ICCV_2017_paper.html}
}

@inproceedings{zhang2019shift,
  title={Making Convolutional Networks Shift-Invariant Again},
  author={Zhang, Richard},
  booktitle={Proceedings of the 36th International Conference on Machine Learning},
  series={Proceedings of Machine Learning Research},
  volume={97},
  pages={7324--7334},
  year={2019},
  url={https://proceedings.mlr.press/v97/zhang19a.html}
}

@inproceedings{chaman2021shift,
  title={Truly Shift-Invariant Convolutional Neural Networks},
  author={Chaman, Anadi and Dokmani{\'c}, Ivan},
  booktitle={Proceedings of the IEEE/CVF Conference on Computer Vision and Pattern Recognition},
  pages={3773--3783},
  year={2021},
  url={https://openaccess.thecvf.com/content/CVPR2021/html/Chaman_Truly_Shift-Invariant_Convolutional_Neural_Networks_CVPR_2021_paper.html}
}

@book{smith2011spectral,
  title={Spectral Audio Signal Processing},
  author={Smith, Julius O.},
  publisher={W3K Publishing},
  year={2011},
  url={https://www.dsprelated.com/freebooks/sasp/Polyphase_Decomposition.html}
}
